\documentclass[10pt, conference, letterpaper]{IEEEtran}
\IEEEoverridecommandlockouts

\usepackage{comment}
\usepackage{tabularx}

\usepackage{cite}
\usepackage{amsmath,amssymb,amsfonts}
\usepackage{algorithmic}
\usepackage{graphicx}
\usepackage{textcomp}
\usepackage{xcolor}
\usepackage{multirow}
\usepackage{amssymb}
\usepackage{pifont} 
\usepackage{booktabs}
\usepackage{siunitx}
\usepackage{epstopdf}
\usepackage{xspace}
\usepackage{CJKutf8}
\usepackage{float}
\usepackage{afterpage}
\usepackage{adjustbox}
\usepackage{stfloats}
\usepackage[sort&compress]{natbib}
\bibpunct{[}{]}{,}{n}{}{;}

\usepackage{threeparttable}

\usepackage{pgfplots}
\usepackage{subcaption}
\pgfplotsset{compat=1.18}
\usepackage{tikz}
\usetikzlibrary{matrix}
\usepackage{tabularx}
\usepackage{comment}
\usepackage[ruled,vlined]{algorithm2e}
\usepackage{amsmath,amssymb}
\usepackage{amsmath}
\usepackage{enumitem}
\usepackage{makecell} 
\usepackage{float}
\usepackage{adjustbox}
\usepackage{graphicx}
\usepackage{array}
\usepackage[utf8]{inputenc}

\usepackage{graphicx}

\usepackage[font=footnotesize]{caption}
\usepackage{subcaption}

\newcommand{\attack}{\textit{$\alpha$-Cloak}\xspace}

\usepackage{tabularx}
\usepackage{array} 

\newcolumntype{Y}{>{\raggedright\arraybackslash\centering\arraybackslash}m{0.4\linewidth}}

\usepackage{array,tabularx} 

\newcolumntype{Y}{>{\centering\arraybackslash}X}

\def\BibTeX{{\rm B\kern-.05em{\sc i\kern-.025em b}\kern-.08em
    T\kern-.1667em\lower.7ex\hbox{E}\kern-.125emX}}
\begin{document}

\title{Can You Trust What You See? Alpha Channel No-Box Attacks on Video Object Detection\\

}

\author{
Ariana Yi$^{1}$ \quad
Ce Zhou$^{2}$ \quad
Liyang Xiao$^{3}$ \quad
Qiben Yan$^{3}$ \\[0.2em]
$^{1}$Mission San Jose High School \quad
$^{2}$Missouri University of Science and Technology \\
$^{3}$Michigan State University \\[0.2em]
\texttt{yiariana7@gmail.com, cezhou@mst.edu, \{xiaoliya, qyan\}@msu.edu}
}

\maketitle

\begin{abstract}

As object detection models are increasingly deployed in cyber-physical systems such as autonomous vehicles (AVs) and surveillance platforms, ensuring their security against adversarial threats is essential. While prior work has explored adversarial attacks in the image domain, those attacks in the video domain remain largely unexamined, especially in the no-box setting. In this paper, we present \attack, the first \textit{no-box} adversarial attack 
on object detectors that operates entirely through the alpha channel of RGBA videos. \attack exploits the alpha channel to fuse a malicious target video with a benign video, resulting in a fused video that appears innocuous to human viewers but consistently fools object detectors. Our attack requires no access to model architecture, parameters, or outputs, and introduces no perceptible artifacts. We systematically study the support for alpha channels across common video formats and playback applications, and design a fusion algorithm that ensures visual stealth and compatibility. We evaluate \attack on five state-of-the-art object detectors, a vision-language model, and a multimodal large language model (Gemini-2.0-Flash), demonstrating a 100\% attack success rate across all scenarios. Our findings reveal a previously unexplored vulnerability in video-based perception systems, highlighting the urgent need for defenses that account for the alpha channel in adversarial settings.

\end{abstract}

\begin{IEEEkeywords}
Video attack, No-box attack, Object detection, LLM security
\end{IEEEkeywords}

\section{Introduction}

Artificial intelligence (AI) models are increasingly integrated into cyber-physical systems, empowering tasks such as obstacle avoidance in autonomous driving, environmental sensing in smart homes, and intelligent motion control in robotics. With the rapid advancement of AI, large language models (LLMs) are also being adopted in these domains~\cite{zhou2024comprehensive}. Due to their low cost and portability, camera sensors have become one of the most widely used sensing modalities in such systems. As a result, computer vision tasks involving both images and videos play a critical role in system functionality.

Despite these advancements, security vulnerabilities persist due to the inherent weaknesses of AI models. To address practical and generalizable threats, black-box adversarial attacks have been widely studied~\cite{andriushchenko2020square,chen2020hopskipjumpattack,moon2019parsimonious,wang2023beyond}. However, existing black-box attacks often suffer from key limitations, including excessive query requirements, low efficiency, and reduced success rates and confidence levels~\cite{xia2025alphadog}. While some black-box attacks have been extended to the physical world~\cite{zhou2022doublestar,zhou2024optical}, they remain constrained by real-world challenges such as physical access and continuous control. Recently, Xia et al.~\cite{xia2025alphadog} proposed AlphaDog, a no-box universal attack, which exploits the previously overlooked alpha channel in images to achieve 100\% success rate and confidence with high stealth. However, their work focuses solely on the image domain, leaving the video domain unexplored.

Notably, various video formats, such as Apple ProRes 4444 (.mov), HEVC (.hevc), WebM (.webm), OpenEXR (.exr), and Animated PNG (.apng), also support alpha channels. In the video domain, the alpha channel functions similarly to that in images. It works as a transparent layer enabling seamless blending of visual elements. It plays a critical role in video editing, web development, and graphic design. In this paper, inspired by AlphaDog~\cite{xia2025alphadog}, we propose the first no-box adversarial attack in the video domain, called \attack, which targets object detection systems commonly deployed in cyber-physical environments.

Unlike the image domain, the video domain presents two unique challenges. First, most video players use black or gray backgrounds by default, rather than white, and not all video formats support alpha channels. To address this, we conduct extensive preliminary experiments to identify compatible video formats and analyze the background colors used by popular video players. Second, embedding an adversarial image into a video is non-trivial because videos consist of multiple frames, not a single static image. To overcome this, we design a novel fusion algorithm that combines a benign video and a malicious adversarial video. By carefully tuning key parameters, our method ensures the malicious content remains completely invisible to human observers while still being detected by AI models.

We evaluate \attack on five widely used object detection models, one vision-language model (VLM), and extend our analysis to an LLM with visual capabilities. Because the adversarial content is embedded structurally within the video format, the attack remains robust across diverse models and video players, achieving a 100\% success rate in all experiments.

Our contributions are summarized as follows:
\begin{itemize}
  \item We present \attack, the first no-box adversarial attack on object detection models processing video inputs. It requires no model queries, architecture knowledge, parameter access, or output feedback during generation.
  \item We demonstrate that adversarially perturbed videos can cause object detection models to consistently perceive a target malicious video while human viewers see only the original benign content. Unlike perturbation-based methods, \attack introduces no visible noise for humans or detectors. 
  \item We validate our approach across a broad range of vision and language models, including various types of object detection models, a VLM, and a  multimodal LLM (Gemini-2.0-Flash). We achieve a 100\% attack success rate across all cases. This highlights the broad applicability and robustness of our attack across diverse architectures and modalities.
\end{itemize}
\section{Background}
In this section, we present background information on the RGBA video format and the role of the alpha channel. We then describe how such videos are rendered by video playback applications and processed by models.

\subsection{RGBA Video Format and Alpha Channel}
Digital videos have pixel formats that define the color and transparency information for each frame. Two pixel formats of the RGB color model are RGB and RGBA. The RGB format stores red, green, and blue channels that determine the color of each pixel. In an RGB video, each pixel stores values for these three channels, enabling the rendering of a wide spectrum of colors. RGBA extends this format by adding a fourth channel, the alpha channel.

\begin{figure*}[!t]
    \centering
    \includegraphics[width=0.65 \textwidth]{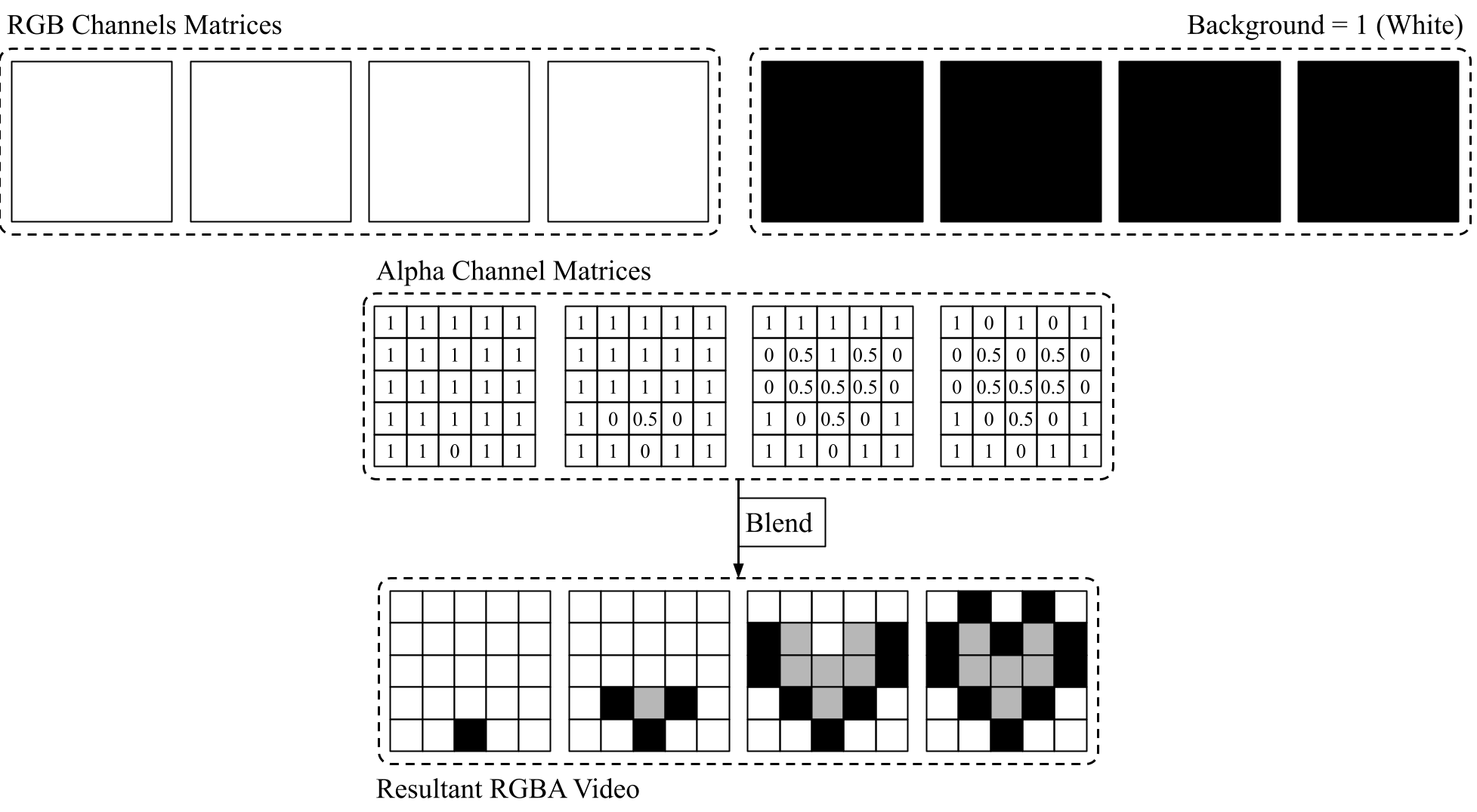}
    \caption{An example of alpha channel blending illustrates how the alpha value interacts with the RGB channels and the background color to control pixel transparency and the final color seen in video display applications.}
    \label{fig:alpha_channel}
\end{figure*}

The alpha channel represents the transparency level of each pixel and typically ranges from 0 to 255 in 8-bit encoding. A value of 255 indicates full opacity, while a value of 0 denotes complete transparency. Intermediate values result in varying degrees of partial transparency. In this paper, we normalize the alpha channel values to a range between 0 and 1 for consistency in the blending calculations. When an RGBA video is rendered, pixels with alpha values less than 1 reveal the underlying background of the video player, with the RGB values blended as an overlay. The final color of each pixel displayed to viewers is computed through an element-wise alpha compositing operation, combining the original RGB values with the background color of the video player.

Formally, for each pixel in the rendered video, the resulting 3-element vector {P} is calculated as a weighted combination of the pixel’s original RGB values {$C_{\text{RGB}}$} and the background color {$C_{\text{BG}}$}, using the normalized alpha value $\alpha \in [0, 1]$ as the blending factor. The compositing formula is:
\begin{equation}
    P = \alpha\cdot {C_{\text{RGB}}} + (1-\alpha) \cdot C_{\text{BG}}.
\end{equation}
This operation is performed independently for each pixel in the frame, therefore applying the formula in an element-wise manner across the entire image. When this equation is applied to all pixels in a frame, this operation produces a 3-D output matrix of size $H \times W \times3$, where {H} and {W} denote the height and width of the video frame, respectively. 

An example of this process is shown in Fig.~\ref{fig:alpha_channel}, where four RGBA video frames are composited over a black background to produce the final visible result. The appearance of the same RGBA video can vary depending on the alpha channel values and the background color rendered by the video player. In this example, each 5×5 square represents a single frame from the video. A group of four such frames together forms a segment of the overall video, demonstrating how alpha blending operates consistently across consecutive frames. 

\subsection{Background Colors of Video Player Applications}

Digital video content can be rendered through various playback environments, including standalone video player applications and embedded viewers within web browsers. These players differ in their support for alpha channels. While some video players can interpret and render the alpha channel, they typically default to displaying the video over a solid background color. As a result, semi-transparent regions in the video may blend with the background, making portions of the background color visible to human viewers. This inconsistency in background handling leads to visual differences in how the same RGBA video appears across different platforms.

In addition to full video playback, most systems also generate video thumbnails, which are small preview images commonly shown in file explorers or gallery applications. These thumbnails are often rendered using a different default background color than the one used during full playback. Table~\ref{bckgd_colors} summarizes the background colors applied by popular video player applications during video playback and thumbnail rendering. The results indicate that \emph{most video players default to a black background during playback, while thumbnail backgrounds tend to alternate between black and gray}.

\begin{table}[t]
\centering
\caption{Background colors of video players.}
\renewcommand{\arraystretch}{1.25}
\setlength{\tabcolsep}{3.45pt} 

\begin{tabularx}{\columnwidth}{|
    >{\raggedright\arraybackslash}m{0.21\columnwidth}|
    >{\raggedright\arraybackslash}m{0.35\columnwidth}|
    >{\raggedright\arraybackslash}m{0.35\columnwidth}|}
  \hline
  \textbf{Background Color} &
  \textbf{Thumbnail (Reduced-Size Video Display)} &
  \textbf{Viewer (Full-Size Video Display)} \\
  \hline
  Black Background &
  VLC Media Player, macOS Finder, Apple TV, Adobe Premiere Pro, Capcut &
  VLC Media Player, QuickTime Player, Apple TV, Microsoft ClipChamp, Adobe Premiere Pro, Capcut, Vimeo Player \\
  \hline
  Grey Background &
  YouTube Player, Google Drive Video Player, OneDrive Player, Amazon Drive, iPhone Photos &
  YouTube Player, Google Drive Video Player, OneDrive Player, Amazon Drive \\
  \hline
  White Background &
  Vimeo Player &
  iPhone Photos \\
  \hline
\end{tabularx}
\label{bckgd_colors}
\end{table}

\subsection{Alpha Channel in Video File Types}

Many video formats, such as .mp4, are designed for RGB videos and do not support alpha channels. Attempting to store transparency in these formats will result in the alpha channel data being discarded, and the video will default to being fully opaque. To retain the alpha channel, the video must be encoded using a file type that explicitly supports the RGBA format. Table~\ref{video_types} outlines some of the most widely used video file formats and their ability to support an alpha channel when displaying videos.

\begin{table}[t]
\centering
\caption{Alpha channel support in various video file formats.}
\renewcommand{\arraystretch}{1.25}
\setlength{\tabcolsep}{2pt} 

\begin{tabularx}{\columnwidth}{|
  >{\raggedright\arraybackslash}m{0.23\columnwidth}|
  >{\raggedright\arraybackslash}m{0.36\columnwidth}|
  >{\raggedright\arraybackslash}m{0.36\columnwidth}|}
  \hline
  & \textbf{Supports Alpha Channel} & \textbf{Does Not Support Alpha Channel} \\ \hline

  \textbf{\makecell{Video File \\ Format \\ (Media Type)}} &
  Apple ProRes 4444 (.mov), HEVC (.hevc), WebM (.webm), OpenEXR (.exr), Animated PNG (.apng) &
  MPEG-4 (.mp4), Audio Video Interleave (.avi), Windows Media Video (.wmv) \\
  \hline
\end{tabularx}
\label{video_types}
\end{table}

\subsection{RGBA Video-Based Object Detectors Processing Pipeline}
Most modern object detectors accept only three‐channel RGB inputs, since they are trained on datasets of standard RGB images or videos~\cite{zou2023object}. When an RGBA video is provided, the alpha channel is removed during preprocessing, either explicitly removed or by converting the input to RGB format, so any transparency information is lost before inference~\cite{xia2025alphadog,Suri2016}.

Modern object detection models are typically categorized into one-stage and two-stage architectures. One-stage detectors, such as YOLOv5 and its successors, perform object classification and localization in a single network pass~\cite{khanam2024yolov5deeplookinternal}. In contrast, two‐stage detectors such as Faster R-CNN~\cite{ren2016faster} first generate region proposals and then classify and refine these proposals in a second network stage. 

Recent advances in VLMs and LLMs have broadened the scope of visual recognition. Models such as Open-VCLIP extend the CLIP framework to videos, learning aligned embeddings between video inputs and textual class labels~\cite{weng2023open}. Gemini-2.0-Flash is a variant of Google DeepMind's multimodal LLM, which processes both text and visual inputs through multimodal embeddings~\cite{team2023gemini}. Both models operate on RGB inputs, with alpha channels ignored or discarded during preprocessing. 
\section{Threat Model}

\begin{figure}
    \centering
    \includegraphics[width=1\linewidth]{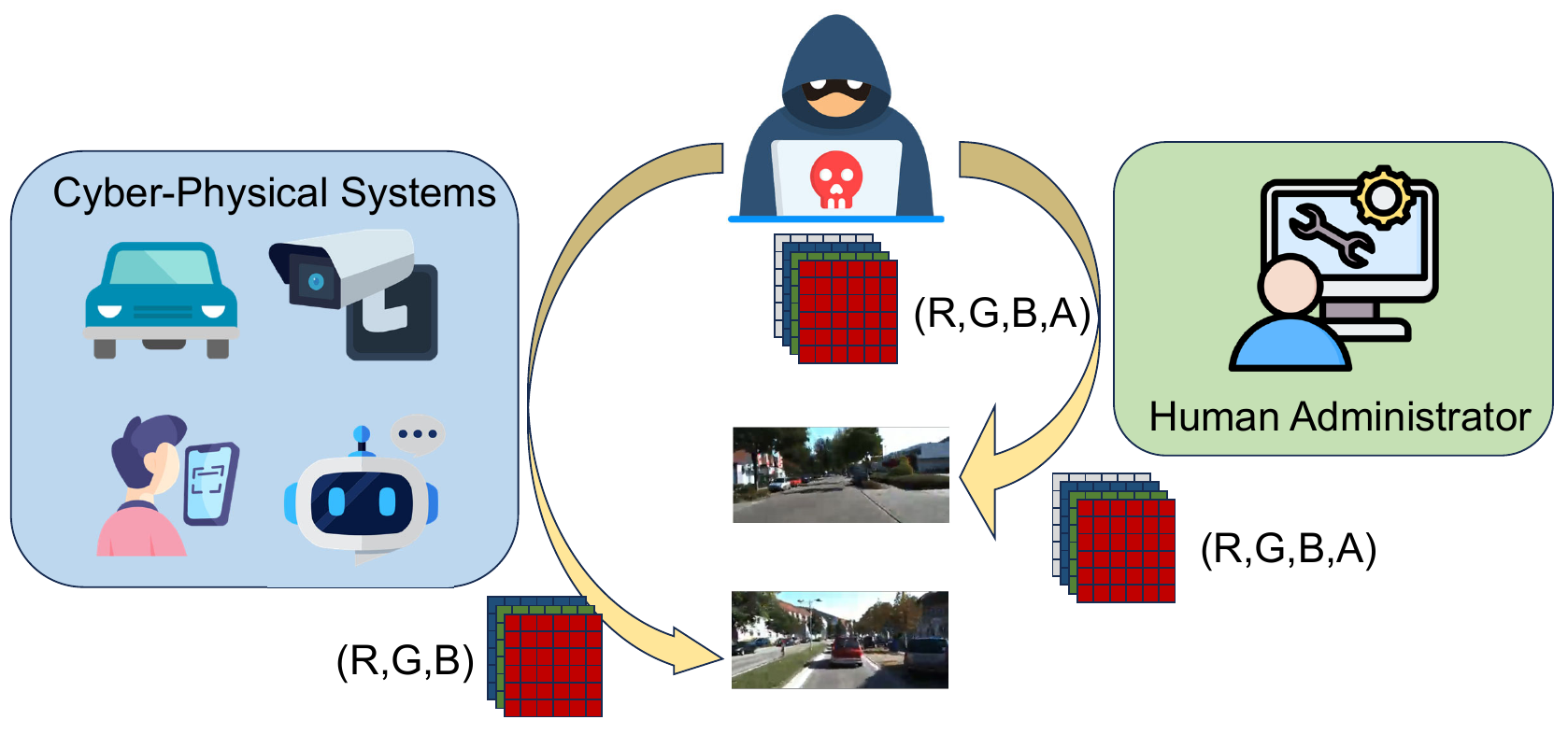}
    \caption{Attack scenario}
    \label{fig:Threat Model}
    \vspace{-15pt}
\end{figure}

As shown in Fig.~\ref{fig:Threat Model}, we consider a threat model grounded in cyber-physical systems where visual perception plays a critical role in system behavior. These systems, including autonomous vehicles (AVs), surveillance systems, face recognition systems and smart home robots, such as Tesla's Full Self-Driving System(FSD)~\cite{tesla_fsd}, mobile robots, such as Starship Technologies' autonomy robots~\cite{starship}, and AI-driven sensing platforms, highly rely on object detectors to interpret the surrounding environment and later make real-time decisions. These models often process video inputs under the assumption that the video input is benign and clean. However, our attack exposes a possible vulnerability for the model when given an RGBA \attack video.

\textbf{Attack Goal.} The attacker’s goal is to hide information in the alpha channel of an RGBA video, and then take advantage of the detector’s preprocessing step that drops this channel. Once the alpha channel is removed, the detector only sees the RGB image, which shows a scene picked by the attacker. This change is invisible to a person but can cause serious safety problems in real systems. For example, an AV using such an attacked video might miss important objects that are not in the attacker’s scene, leading to wrong and potentially dangerous driving decisions.

\textbf{Attacker's Capabilities.} \attack is a no‑box attack that requires no knowledge of the detection model’s parameters or architecture. In our threat model, the attacker aims at compromising the object detector of a cyber‑physical system. To launch the attack, they craft an adversarial RGBA \attack video by merging two streams: a malicious “target” video meant for the detector and a benign video meant for human viewers. Because the detector accepts only RGB inputs and drops the alpha channel, it processes only the attacker’s chosen scene. In addition, the attacker never needs physical access to the device; the attacked video can be delivered digitally (for example, via standard media uploads).

\textbf{Attack Scenarios.} Fig.~\ref{fig:Threat Model} illustrates the attack workflow in a cyber‑physical system. The adversary first creates an RGBA video by embedding a malicious target RGB stream into the alpha channel of a benign RGB video. This single tampered file is then uploaded through the cloud-based media interface of the system to the human administrator and the vehicle perception module. Because the perception module drops the alpha channel, it processes only the attacker’s chosen frames, while the human operator sees the benign original sequence. Consequently, an AV may interpret a busy road as empty, leading to unsafe actions such as unintended lane changes or sudden acceleration.

\section{\attack Attack Design}

In this section, we present the attack design of \attack. We first present an overview of the attack, and then we detail how the attack is conducted on each frame of the video. 

\subsection{Attack Overview}

\begin{figure*}[t]
    \centering
    \includegraphics[width=0.7 \textwidth]{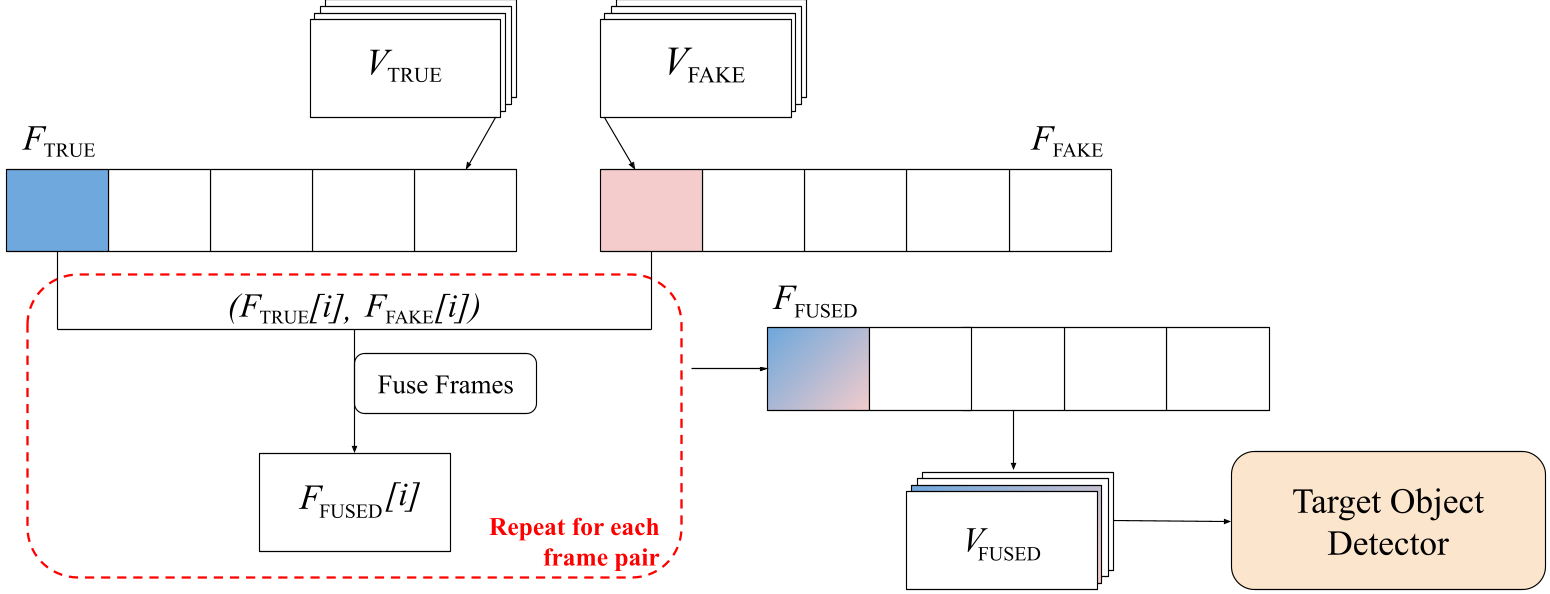}
    \caption{Overview of the \attack pipeline}
    \label{fig:attack_design}
\end{figure*}

As shown in Fig.~\ref{fig:attack_design}, we define three video streams in \attack: \( V_{\text{TRUE}} \), \( V_{\text{FAKE}} \), and \( V_{\text{FUSED}} \).  
\( V_{\text{TRUE}} \) represents the benign, human-visible video, while \( V_{\text{FAKE}} \) denotes the malicious target video intended to deceive AI perception systems. The final adversarial output, \( V_{\text{FUSED}} \), is generated by fusing \( V_{\text{TRUE}} \) and \( V_{\text{FAKE}} \) in a manner that preserves visual normality to human viewers but induces incorrect outputs in object detection models.

To construct \( V_{\text{FUSED}} \), both \( V_{\text{TRUE}} \) and \( V_{\text{FAKE}} \) are first decomposed into their constituent frames. Let \( F_{\text{TRUE}} \) and \( F_{\text{FAKE}} \) denote the respective arrays of frames. For each corresponding pair \( (F_{\text{TRUE}}[i], F_{\text{FAKE}}[i]) \), we apply the \textsc{FuseFrames} method to produce a fused frame, and further resulting in the array \( F_{\text{FUSED}} \). 
The fused frames are then reassembled into a video at a consistent frame rate to produce the final adversarial video \( V_{\text{FUSED}} \), which appears visually identical to \( V_{\text{TRUE}} \) to human observers but is interpreted by detection models as \( V_{\text{FAKE}} \). The full fusion pipeline is described in Algorithm~\ref{alg_1}.

\begin{algorithm}[htbp!]
\LinesNumbered
\caption{Generate \attack Fused Video} \label{alg_1}
\SetKwInOut{Input}{Input}
\SetKwInOut{Output}{Output}
\SetKwProg{Fn}{function}{:}{}

\Input{Benign Video $V_{\text{TRUE}}$, Malicious Video $V_{\text{FAKE}}$, Frame Width $l$, Frame Height $h$}
\Output{The generated attack video $V_{\text{FUSED}}$}

\Fn{\textsc{GenerateFusedVideo}($V_{\text{TRUE}}, V_{\text{FAKE}}, l, h$)}{
    $F_{\text{TRUE}} \gets \textsc{FramePrep}(V_{\text{TRUE}}, l, h)$\;
    $F_{\text{FAKE}} \gets \textsc{FramePrep}(V_{\text{FAKE}}, l, h)$\;
    
    \For{$i \gets 0$ \KwTo $\min(\textnormal{len}(F_{\text{TRUE}}), \textnormal{len}(F_{\text{FAKE}}))$}{
        $F_{\text{FUSED}}[i] \gets \textsc{FuseFrames}(F_{\text{TRUE}}[i], F_{\text{FAKE}}[i])$\;
    }
    
    $V_{\text{FUSED}} \gets \textsc{generate\_video}(F_{\text{FUSED}})$\;
    \textbf{return} $V_{\text{FUSED}}$\;
}

\Fn{\textsc{FramePrep}($V$, $l$, $h$)}{
    $V \gets \textsc{resize}(V, l, h)$\;
    $F \gets \textsc{split\_into\_frames}(V) \rightarrow \{F_1, F_2, \ldots, F_n\}$\;
    \textbf{return} $F$\;
}

\end{algorithm}

\subsection{Video Frame Preprocessing}

To ensure successful fusion between the two input videos \( V_{\text{TRUE}} \) and \( V_{\text{FAKE}} \), we first preprocess the two videos using the \textsc{FramePrep} function outlined in Algorithm~\ref{alg_1}. The purpose of this step is to standardize spatial and temporal properties between inputs and to allow frame-level access for subsequent functions. The input videos are rescaled to a uniform frame size \( l \times h \), ensuring compatibility for pixel-wise fusion. Finally, we split each video into an array of its frames, which enables direct access and manipulation during the fusion process.

\begin{algorithm}[htbp!]
\LinesNumbered
\caption{Generate Fused Frame} \label{alg_2}
\SetKwInOut{Input}{Input}
\SetKwInOut{Output}{Output}
\SetKwProg{Fn}{function}{:}{}

\Input{Benign Frame $F_{\text{TRUE}}$, Malicious Frame $F_{\text{FAKE}}$.}
\Output{The generated fused attack frame $V_{\text{FUSED}}$.}

\Fn{\textsc{FuseFrames}($F_{\text{TRUE}}, F_{\text{FAKE}}$)}{
    $F_{\text{TRUE}} = \textsc{Preprocess} (F_{\text{TRUE}}) \times 0.4$\;
    
    $F_{\text{FAKE}} = \textsc{Preprocess} (F_{\text{FAKE}}) \times 0.6 + 0.4$\;
    
    \vspace{1em}
    
    $A_{\text{FUSED}} =  \frac{F_{TRUE}}{F_{FAKE}}$
    \vspace{1em}

    $F_{\text{FUSED}} = \textnormal{concatenate}(A_{\text{FUSED}}, F_{\text{FAKE}})$\; 

    $F_{\text{FUSED}} = F_{\text{FUSED}} \times 255.0$\;

    \textbf{return} $F_{\text{FUSED}}$\;
}

\Fn{\textsc{Preprocess}($F$)}{
    $F$ = grayscale($F$)
    
    $F = F \div 255.0$\;

    \textbf{return} $F$;
}

\end{algorithm}

\subsection{Video Frames Combination}

We perform frame fusion between both arrays of frames to generate a single composite video that embeds information from both \( V_{\text{TRUE}} \) and \( V_{\text{FAKE}} \). After splitting the input videos \( V_{\text{TRUE}} \) and \( V_{\text{FAKE}} \) into frame arrays $F_{\text{TRUE}}$ and $F_{\text{FAKE}}$, we apply the \textsc{FuseFrames} function to each corresponding frame pair ($F_{\text{TRUE}}$[$i$], $F_{\text{FAKE}}$[$i$]). This per-frame fusion step is shown in lines 5 and 6 of Algorithm~\ref{alg_2}.

To prepare the frames for fusion, we convert them both to grayscale and normalize their pixel intensities. We normalize pixel values to the range [0,1] by dividing each pixel by its maximum intensity value, i.e., 255 in 8-bit images. 

To ensure that the content of $F_{\text{TRUE}}[i]$ and $F_{\text{FAKE}}[i]$ each remain perceptible in the fused output to their intended targets, while remaining imperceptible to the unintended side, we constrain the intensity ranges of both input frames. We adjust the frames such that $F_{\text{TRUE}} \leq F_{\text{FAKE}}$, ensuring that the alpha channel remains within the normalized bounds [0,1]. We calculate the alpha channel matrix $A_{\text{FUSED}}[i]$ using the following formula: 
\begin{equation}
    A_{FUSED}[i] = \frac{F_{TRUE}[i]}{F_{FAKE}[i]}
    \label{equ1}.
\end{equation}
Substituting Equation (\ref{equ1}) into the inequality, we have the following:
\begin{equation}
    0 \leq \frac{F_{TRUE}[i]}{F_{FAKE}[i]} \leq 1,
\end{equation}
which directly implies the constrait $F_{\text{TRUE}} \leq F_{\text{FAKE}}$ across all pixels.

We empirically determine an optimal intensity range for both input videos. Through experimentation with 6,680 generated \attack videos, we find that scaling $F_{\text{TRUE}}$ to 40\% of its original intensity, while maintaining $F_{\text{FAKE}}$ values above 0.4 achieves the highest performing fusion quality. This results in the following bound:
\begin{equation}
    0 \leq {F_{TRUE}[i]} \leq 0.4 \leq {F_{FAKE}[i]} \leq 1.
\end{equation}

We finalize each fused frame by combining the computed alpha channel matrix with its RGB channel intensity matrix. Because systems typically remove the alpha channel matrix when rendering the frame, we assign the RGB channel intensity matrix equal to $F_{\text{FAKE}}$, as this will be the only image that the computer will see. Thus, the resulting fused frame is: \( F_{\text{FUSED}}[i] = A_{\text{FUSED}}[i] + V_{\text{FAKE}}[i] \).

\section{Evaluation}

To evaluate the proposed attack, we input adversarially fused videos into multiple object detection models and measure how closely their predictions align with the content of either the benign or malicious source video.

\subsection{Experimental Setup}

\subsubsection{Experimental Procedure.}
We design this experiment to evaluate the extent to which object detection models can identify and localize objects within adversarially fused videos. Each target model receives a list of attacked videos along with the ground truth bounding box labels corresponding to each original, unaltered input video used to construct those attacked videos. 

For each frame in a given attacked video, the detection model performs inference and outputs its predicted bounding boxes. We then compute a frame-level similarity score (\textit{FLS}) by comparing the model prediction to all ground truth boxes. This process is repeated for each frame, and the resulting \textit{FLS} values are averaged to obtain a video-level similarity score (\textit{VLS}) between the attacked video and each candidate source video. The candidate source video with the highest average similarity is identified as the most likely source video, indicating that the model's predictions most closely resemble that video's object layout. This experiment allows us to quantify how closely the fused content influences model perception and how effectively the attack obscures source attribution.

\subsubsection{Target Attack Models.} 
We evaluate our attack using five widely adopted object detection architectures, selected to cover a diverse range of model structures. Specifically, we test three versions of YOLO, including YOLOv5n~\cite{Jocher_YOLOv5_by_Ultralytics_2020}, YOLOv8n~\cite{yolov8_ultralytics}, and YOLOv11n~\cite{yolo11_ultralytics}, using their official pre-trained weights. Additionally, we include RetinaNet~\cite{lin2018focallossdenseobject} and Faster R-CNN in our evaluation, both of which utilize a ResNet-50 backbone with a Feature Pyramid Network (FPN) to enhance multi-scale feature extraction. 

While our evaluation focuses on standard object detection benchmarks using widely adopted architectures (YOLOv5/8/11, Faster R-CNN, RetinaNet), these models constitute the core perception modules in many modern vision-based systems, including autonomous vehicles, surveillance platforms, and robotics pipelines. Evaluating at this level allows us to precisely measure the model-level effects of our attack, which directly influence downstream system behavior. 

All models generate bounding boxes along with associated class confidence scores for each input video. We apply a fixed confidence threshold of 0.25 across all models to increase object recall, prioritizing detection coverage over precision. This choice ensures that our similarity metric is sensitive to all detectable object instances. This diverse selection of models provides a robust basis for evaluating the generalizability and effectiveness of our attack strategy.

\subsubsection{Evaluation Metrics.}
To assess the similarity between each attacked video and its potential source videos, we introduce a two-level similarity metric framework: \textit{FLS} and \textit{VLS}. These metrics rely on spatial overlap between predicted and ground truth boxes, measured using Intersection over Union (IoU), how much two bounding boxes overlap. 

For each predicted box \textit{p} in a given attacked frame, we compute the IoU against each ground truth box \textit{g} across all candidate videos and retain the maximum value:
\begin{equation}
    \textnormal{IoU}(A,B) = \frac{\textnormal{area}(A\cap B)}{\textnormal{area}(A\cup B)}.
\end{equation}
\begin{equation}
\
FLS = \frac{1}{n}\sum_{i=1}^n \textnormal{max}_{j=1}^m(\textnormal{IoU}(p_i, g_j)).
\end{equation}

We repeat this process for every frame in the attacked video and then compute the average \textit{FLS} across all \textit{T} frames to compute the \textit{VLS} for each candidate video:
\begin{equation}
\
VLS = \frac{1}{T}\sum_{t=1}^T FLS_t.
\end{equation}

This resulting \textit{VLS} for every candidate video captures how closely each candidate video matches the video detected by the object detector. A higher \textit{VLS} indicates stronger alignment between the candidate's ground-truth content and the detector's predictions on the attacked video. These metrics allow us to quantify and assess how convincingly our attack blends the benign and malicious videos.

\begin{table*}[t]
\centering
\caption{Object detectors attack performance on three videos.}
\renewcommand{\arraystretch}{1.5}
\resizebox{\textwidth}{!}{%
\begin{tabular}{|>{\centering\arraybackslash}m{0.3\textwidth} 
                |>{\centering\arraybackslash}m{0.3\textwidth} 
                |>{\centering\arraybackslash}m{0.3\textwidth} 
                |>{\centering\arraybackslash}m{0.15\textwidth}|}
\hline
\textbf{Target frame seen by humans ($V_{TRUE}$)} & 
\textbf{Target frame seen by object detectors ($V_{FAKE}$)} & 
\textbf{Object Detectors Output} & 
\textbf{Object Detector} \\
\hline

\includegraphics[width=\linewidth]{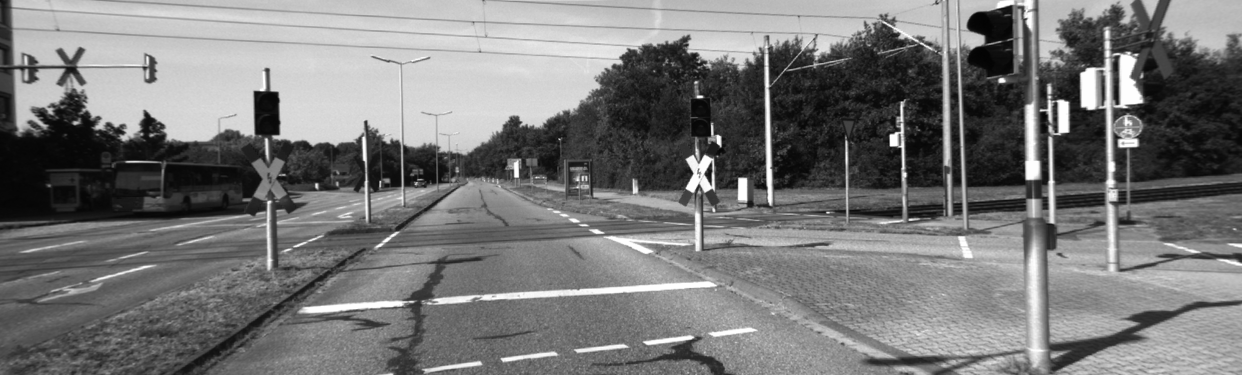} & 
\includegraphics[width=\linewidth]{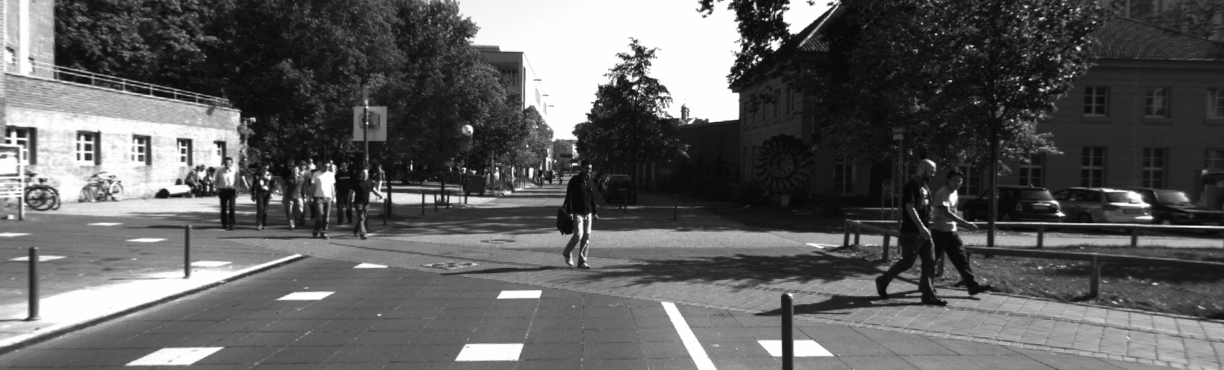} & 
\includegraphics[width=\linewidth]{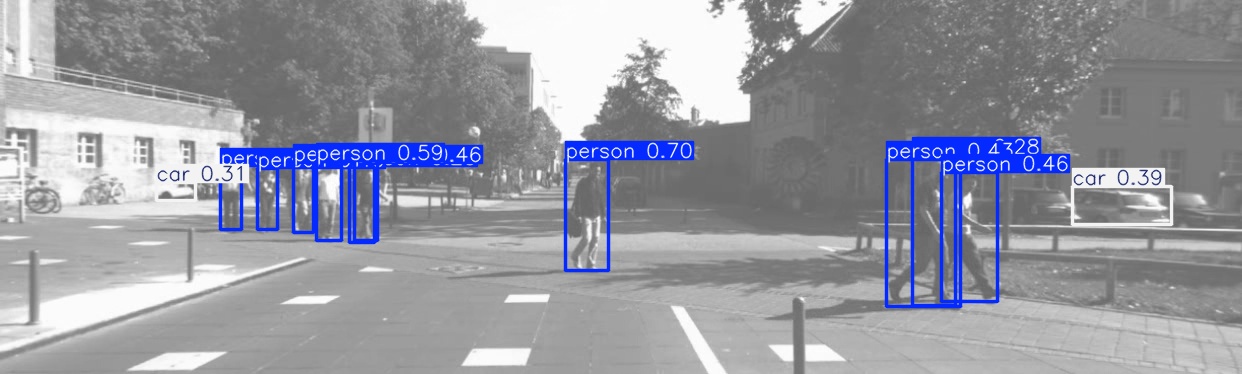} & 
YOLOv5 \\
\hline

\includegraphics[width=\linewidth]{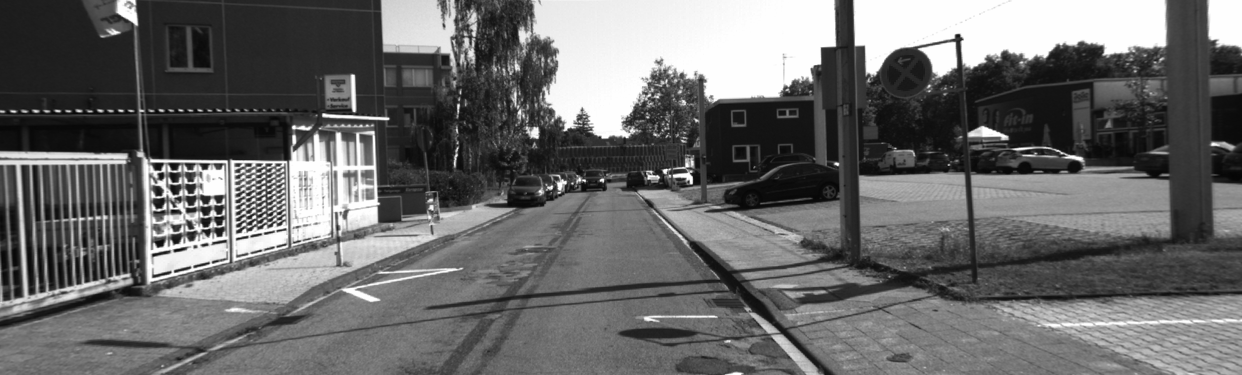} & 
\includegraphics[width=\linewidth]{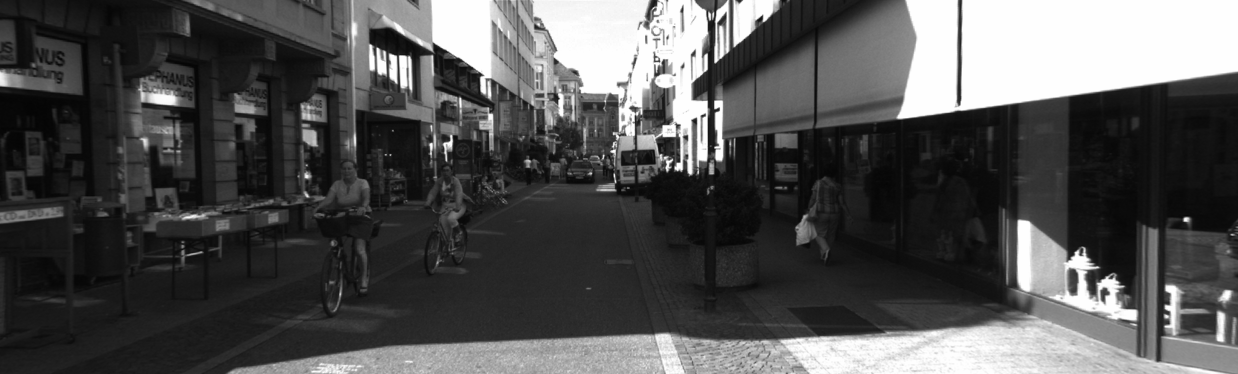} & 
\includegraphics[width=\linewidth]{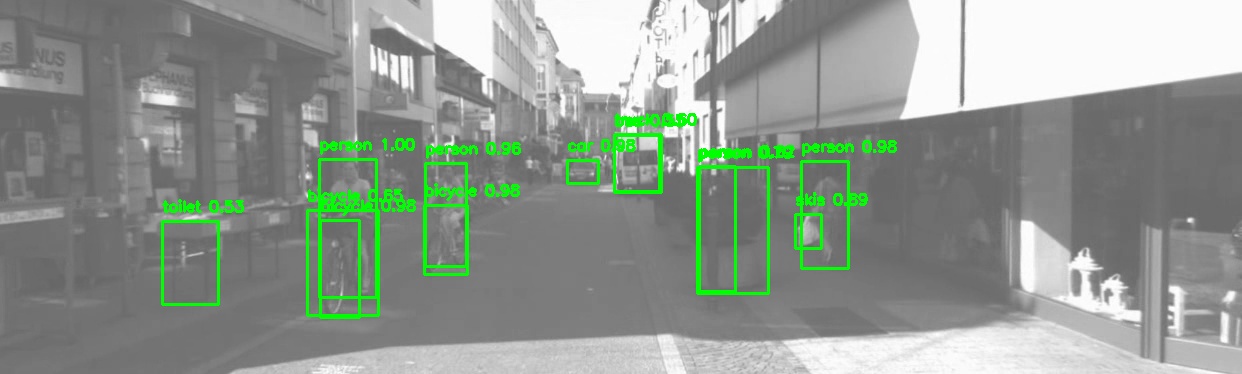} & 
YOLOv11 \\
\hline

\includegraphics[width=\linewidth]{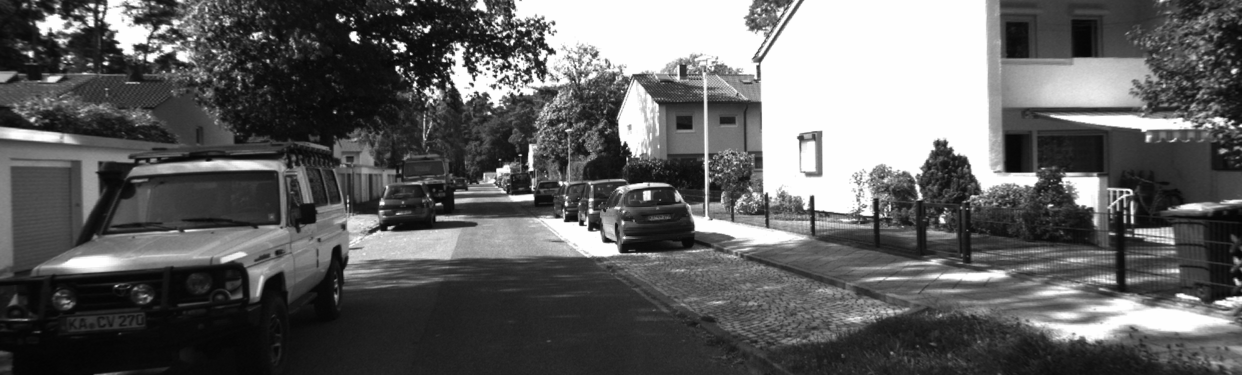} & 
\includegraphics[width=\linewidth]{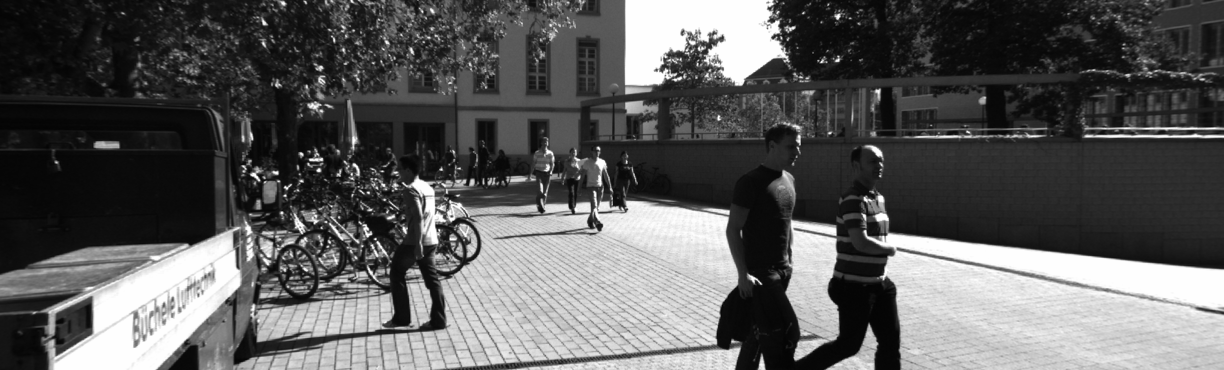} & 
\includegraphics[width=\linewidth]{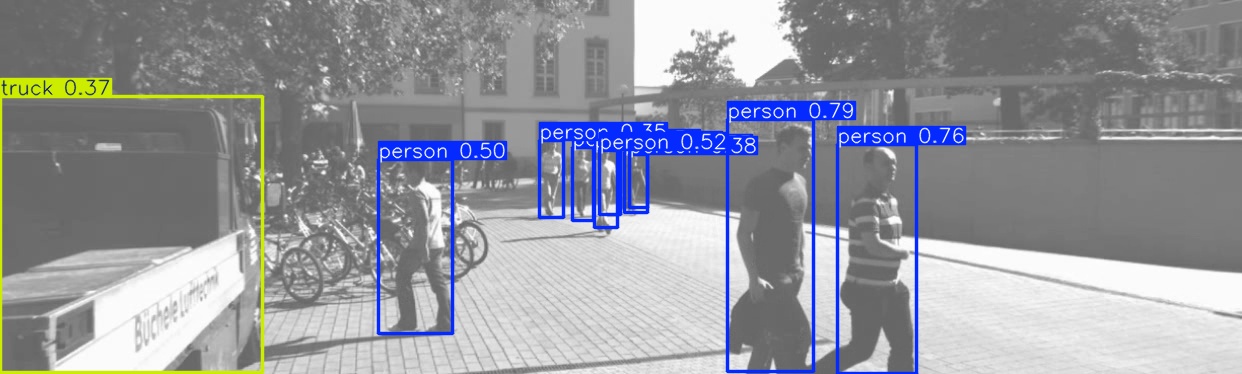} & 
RetinaNet \\
\hline
\end{tabular}%
}
\label{kitti_ex}
\end{table*}

\subsubsection{Dataset.}
We conduct our evaluations using the KITTI tracking dataset~\cite{geiger2013vision}, which provides annotated video sequences for real-world urban driving. We convert the individual KITTI tracking sequences into full-length videos, yielding a set of 21 complete candidate videos. We split the first 20 videos into two equal subsets: the first 10 videos serve as \( V_{\text{TRUE}} \) videos and the next 10 videos as \( V_{\text{FAKE}} \) videos. These fused videos, along with the original KITTI training labels for all 21 videos, are provided as input to the object detection models during evaluation. This experiment design allows us to evaluate attack performance in a multi-object urban context with dynamic backgrounds. Table~\ref{kitti_ex} showcases examples of attacked frames alongside bounding box outputs from three models. The first and second columns are frames that humans and AI should see, respectively. The third and fourth columns show the bounding boxes predicted for each frame, and which model ran the image. It can be seen that the object detector views the \( V_{\text{FAKE}} \) image and runs its object detection on the malicious target video.

\begin{figure}[t]
  \centering
  \begin{subfigure}[b]{0.48\columnwidth}
    \centering\includegraphics[width=\linewidth]{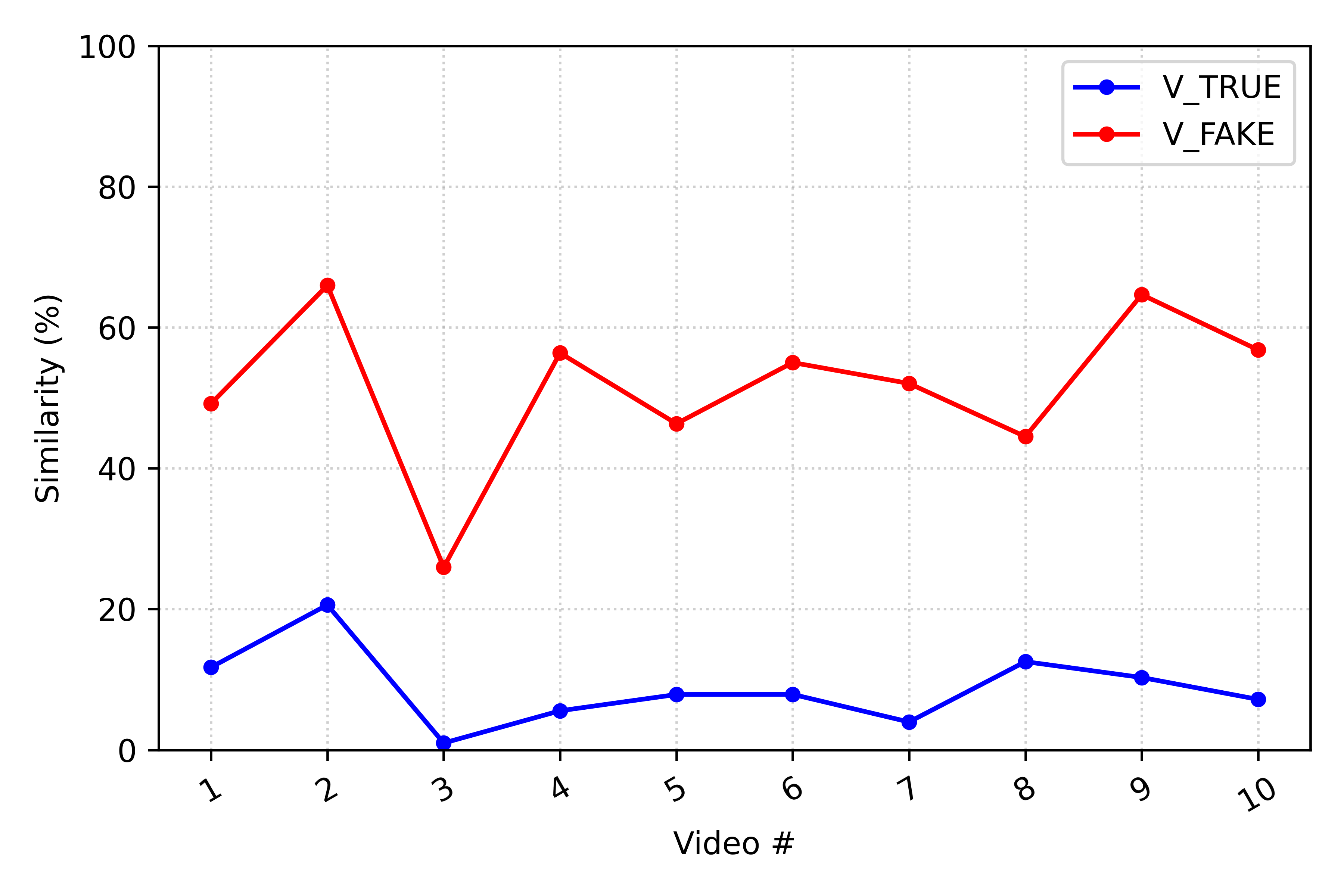}
    \caption{YOLOv5}
  \end{subfigure}\hfill
  \begin{subfigure}[b]{0.48\columnwidth}
    \centering\includegraphics[width=\linewidth]{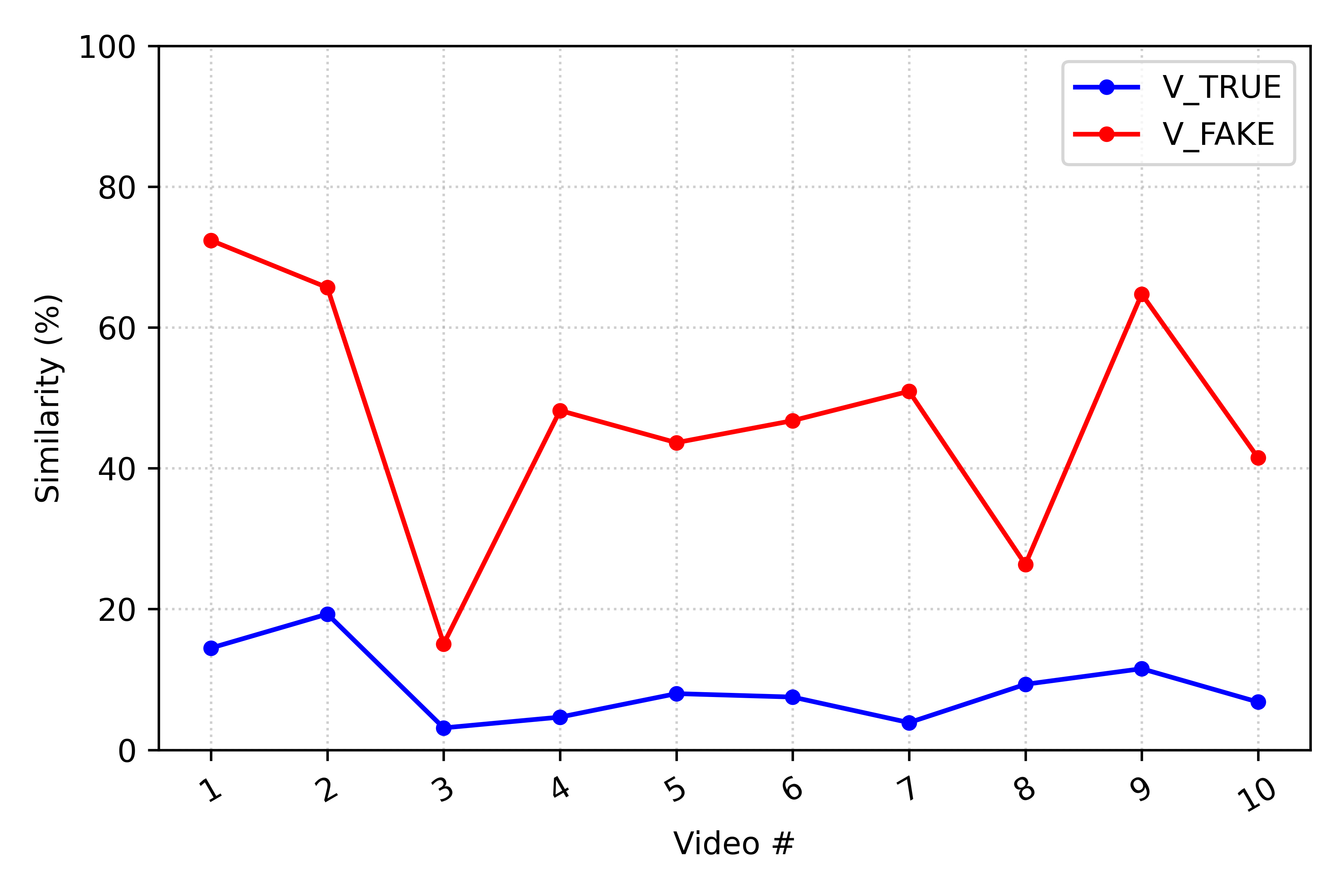}
    \caption{YOLOv8}
  \end{subfigure}

  \begin{subfigure}[b]{0.48\columnwidth}
    \centering\includegraphics[width=\linewidth]{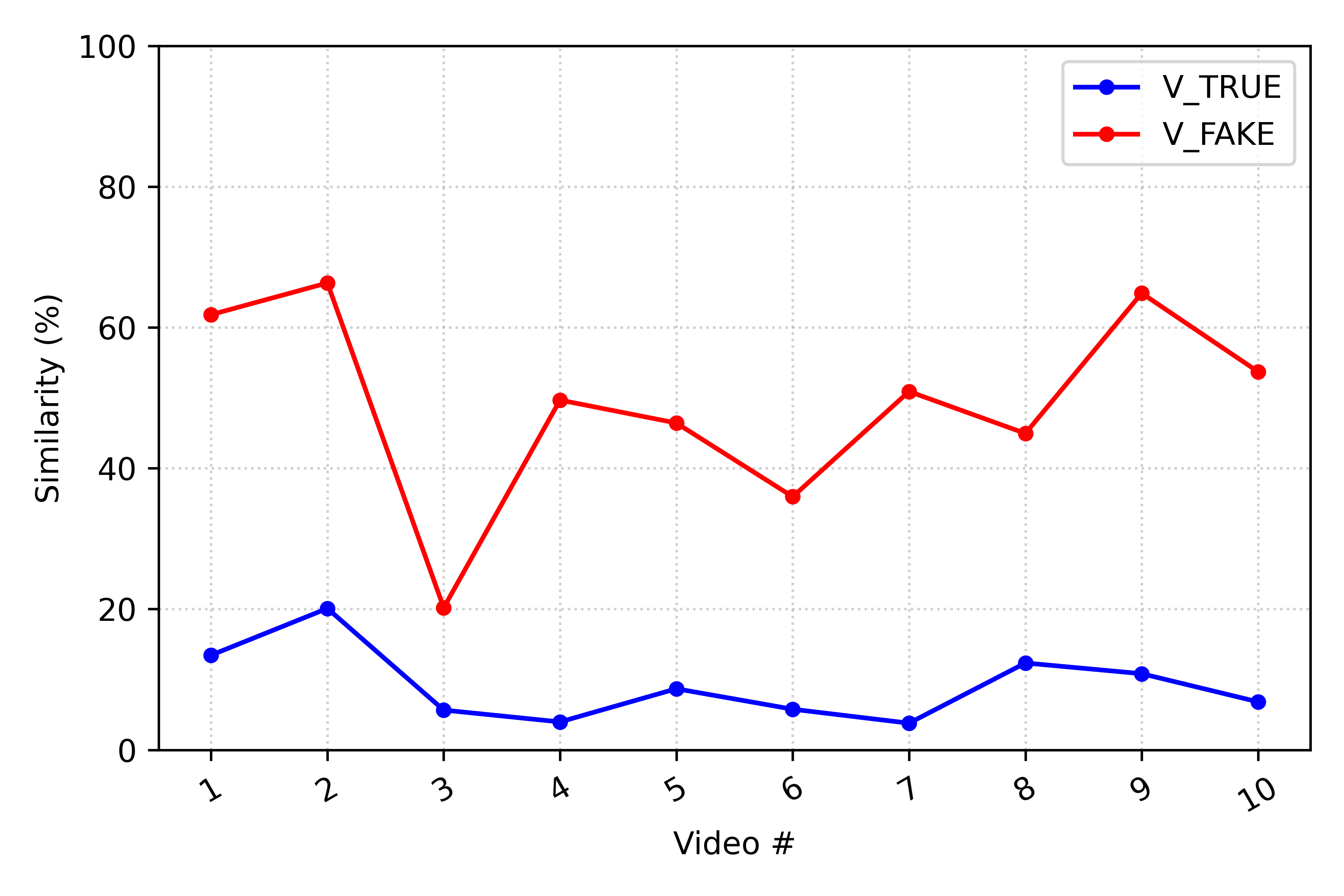}
    \caption{YOLOv11}
  \end{subfigure}\hfill
  \begin{subfigure}[b]{0.48\columnwidth}
    \centering\includegraphics[width=\linewidth]{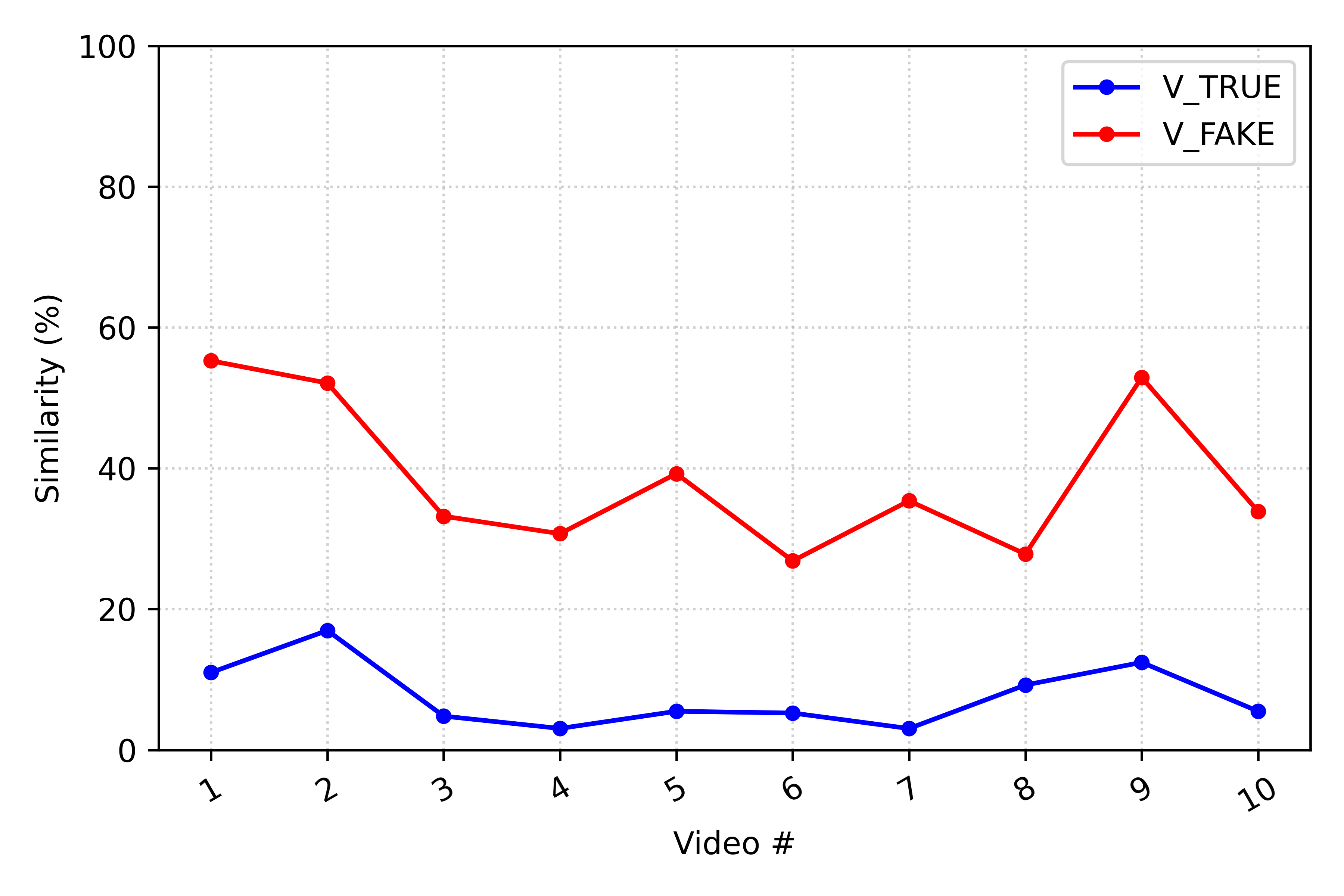}
    \caption{RetinaNet}
  \end{subfigure}

  \begin{subfigure}[b]{0.49\columnwidth}
    \centering\includegraphics[width=\linewidth]{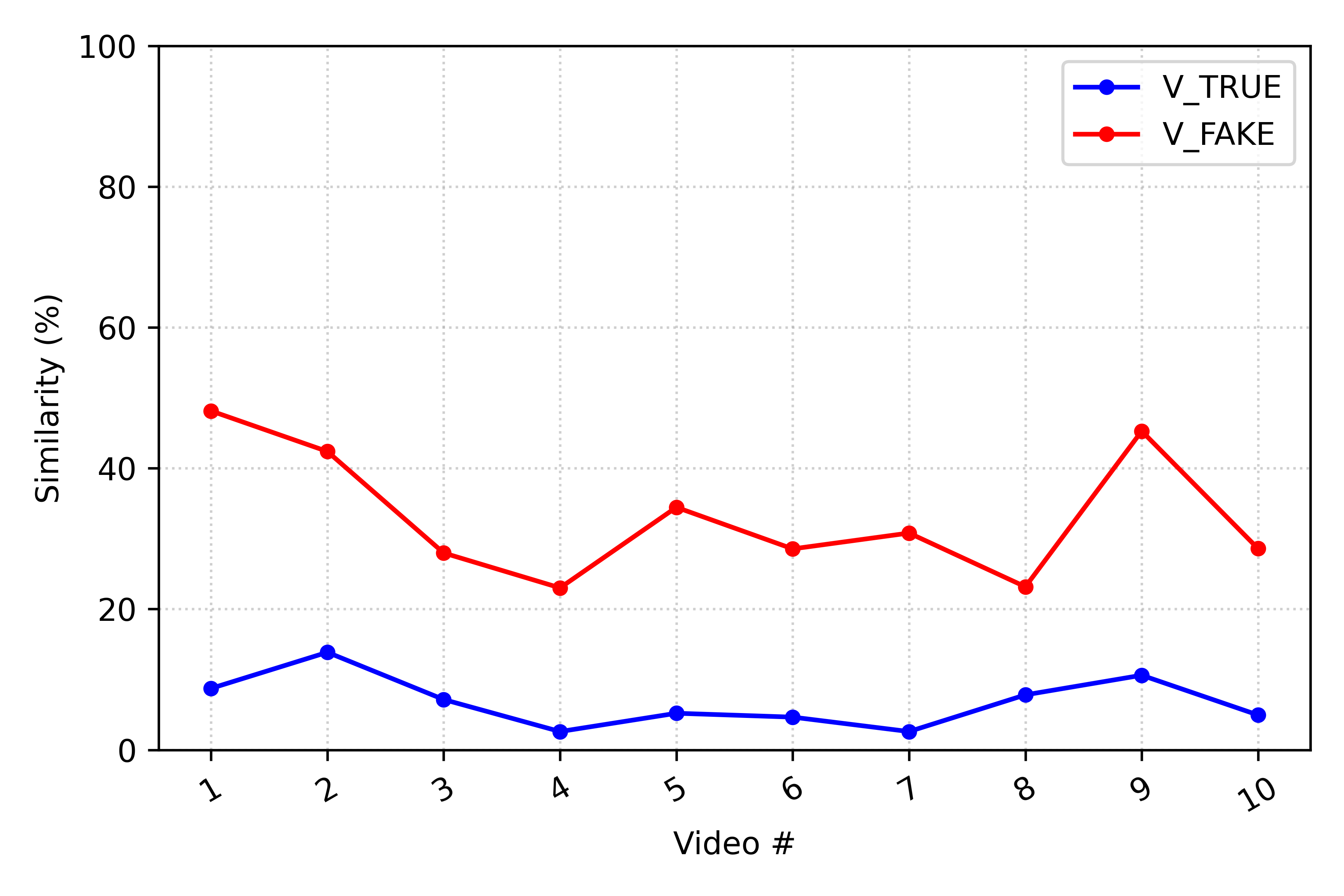}
    \caption{Faster R-CNN}
  \end{subfigure}

  \caption{Evaluation results across different detection models. 
  The x-axis represents each attacked video; the y-axis is the VLS (\%). 
  Red: \(V_{\text{FAKE}}\); blue: \(V_{\text{TRUE}}\).}
  \label{fig:evalKT}
  \vspace{-15pt}
\end{figure}

\subsection{Attack Results on KITTI}
We evaluate the effectiveness of our attack by measuring how closely the fused videos resemble the source components across five object detection models. For each attacked video, we compute the {VLS} relative to its corresponding \( V_{\text{TRUE}} \) and \( V_{\text{FAKE}} \) videos, and plot the results in Fig.~\ref{fig:evalKT}.

\begin{table}[t]
\centering
\caption{Summary of VLS evaluation results across the different object detection models.}
\renewcommand{\arraystretch}{1.25} 
\setlength{\tabcolsep}{3pt}       

\begin{tabularx}{\columnwidth}{|Y|Y|Y|Y|Y|}
\hline
\textbf{Model} &
\textbf{Avg. VLS to $V_{FAKE}$ (\%)} &
\textbf{Avg. VLS to $V_{TRUE}$ (\%)} &
\textbf{$V_{FAKE}$ Top-1 (\%)} &
\textbf{$V_{TRUE}$ Top-1 (\%)} \\ \hline

YOLOv5 & 51.689 & 8.852 & 100 & 0 \\ \hline
YOLOv8 & 47.424 & 8.852 & 100 & 0 \\ \hline
YOLOv11 & 49.487 & 9.138 & 100 & 0 \\ \hline
RetinaNet & 38.733 & 7.665 & 100 & 0 \\ \hline
Faster R-CNN & 33.237 & 6.715 & 100 & 0 \\ \hline
\textbf{OVERALL AVG} & \textbf{44.114} & \textbf{8.244} & \textbf{100} & \textbf{0} \\ \hline
\end{tabularx}
\label{eval_avgs}
\end{table}

Across all models and videos, we observe that the red line, representing the {VLS} of \( V_{\text{FAKE}} \), consistently lies above the blue line, the {VSL} of \( V_{\text{TRUE}} \). This consistent pattern demonstrates that the object detection models perceived the fused videos as more similar to \( V_{\text{FAKE}} \), indicating the success of the attack in misleading the models.

While the difference between the {VLS} scores of \( V_{\text{FAKE}} \) and \( V_{\text{TRUE}} \) are not extreme in magnitude, the direction of the shift is consistent across all models and examples. We attribute the limited gap to the intrinsic visual similarity among the KITTI videos, as all scenes are captured from a vehicle through similar environments. This inherent similarity likely increases the challenge of precise disambiguation.

We summarize quantitative results in Table~\ref{eval_avgs}, which reports the average {VLS} for each model with respect to both source videos, as well as the percentage of cases where \( V_{\text{FAKE}} \) or \( V_{\text{TRUE}} \) was identified as the top-1 most similar candidate. In all models, \( V_{\text{FAKE}} \) leads as the top-1 prediction. This result confirms that our attack effectively redirects model recognition from the benign video and towards the malicious target video, regardless of model architecture.

\subsection{Attack Results on Open-VCLIP}
To evaluate the transferability of our attack beyond object detection, we test it on a VLM, Open-VCLIP, using the UCF-101 dataset~\cite{soomro2012ucf101}. Open-VCLIP is a contrastively trained extension of CLIP designed for video inputs, and it is pre-trained on UCF-101 itself, making it highly familiar with the dataset’s class distribution.

Given a video clip and a predefined set of class labels, Open-VCLIP computes embeddings for both the video input and each class label, ranking all class labels based on cosine similarity, and returns the top-1 nd top-5 predicted classes. This allows us to test whether our attack changes the model's prediction to favor the malicious adversarial label.

We generate 6,660 adversarially fused videos by splitting UCF-101 into two disjoint halves. Videos in the first subset serve as \( V_{\text{TRUE}} \), while the second subset serves as \( V_{\text{FAKE}} \). For each index {i}, we construct a fused video \(V_{\text{FUSED}}^{(i)}\) by applying fusion (Algorithm~\ref{alg_1}) to the pair \(\bigl(V^{(i)}_{\text{TRUE}},\,V^{(i)}_{\text{FAKE}}\bigr)\).  

We submit each \( V_{\text{FUSED}} \) video to Open-VCLIP twice: once with the class label of \( V_{\text{TRUE}} \), and once with the class label of \( V_{\text{FAKE}} \). For both runs, we check whether the submitted label appears in the model's top-1 or top-5 predictions. Table~\ref{eval_ucf} summarizes the classification accuracy under both conditions across all 6,660 attacked videos. The middle row displays the detection accuracy of \( V_{\text{FUSED}} \) using the class label corresponding to the human-visible video \( V_{\text{TRUE}} \). The bottom row displays the detection accuracy of \( V_{\text{FUSED}} \) when using the class label corresponding to the AI-targeted video \( V_{\text{FAKE}} \).

\begin{table}[t]
  \centering
  \caption{Top-1 and Top-5 classification accuracy of Open VCLIP.}
  \renewcommand{\arraystretch}{1.2}
  \setlength{\tabcolsep}{6pt}
  \footnotesize
  \begin{tabular}{|l|c|c|}
    \hline
    \textbf{Label Given} & 
    \textbf{\% of labels in Top-1 list} & 
    \textbf{\% of labels in Top-5 list} \\ \hline

    \(V_{\text{TRUE}}\) Labels & 0.06 & 1.83 \\ \hline
    \(V_{\text{FAKE}}\) Labels & 71.56 & 90.68 \\ \hline
  \end{tabular}
  \label{eval_ucf}
\end{table}

The label of \( V_{\text{TRUE}} \) appears in the top-1 prediction less than 2\% of cases, while the label of \( V_{\text{FAKE}} \) appears in over 90\% of top-1 predictions. This strong skew in prediction distribution indicates that Open-VCLIP consistently aligns with the adversarial target \( V_{\text{FAKE}} \), even in the presence of benign content \( V_{\text{TRUE}} \), highlighting the generalizability and strength of our attack across modalities.

\subsection{\attack Attack Example on Gemini-2.0} 
We extend our investigation on a commercial multimodal LLM, Gemini‑2.0‑Flash. As shown in the Table~\ref{tab:attack_performance_llm}, the first column presents video frames that have been covertly manipulated via the alpha channel, which appear entirely normal to human observers. The second column shows the corresponding attack samples after the large language model strips away the alpha channel. It can be seen that from the LLM’s point of view, these frames convey a completely different scene. The third column reports the model’s analysis of each attack sample. From the results, we can see that the LLM reads the adversarial videos $V_{FAKE}$ instead of $V_{TRUE}$ and demonstrates the effectiveness of \attack on Gemini‑2.0‑Flash.

\begin{table}[t]
  \centering
  \caption{Attack performance on Gemini-2.0-Flash.}
  \renewcommand{\arraystretch}{1.5}

  \begin{adjustbox}{max width=\columnwidth}
    \begin{tabular}{|>{\centering\arraybackslash}m{0.28\textwidth}|
                    >{\centering\arraybackslash}m{0.28\textwidth}|
                    >{\arraybackslash}m{0.38\textwidth}|}
      \hline
      \textbf{Target frame seen by humans ($V_{TRUE}$)} &
      \textbf{Target frame seen by object detectors ($V_{FAKE}$)} &
      \textbf{Gemini-2.0-Flash Output} \\ \hline

      \includegraphics[width=\linewidth,keepaspectratio]{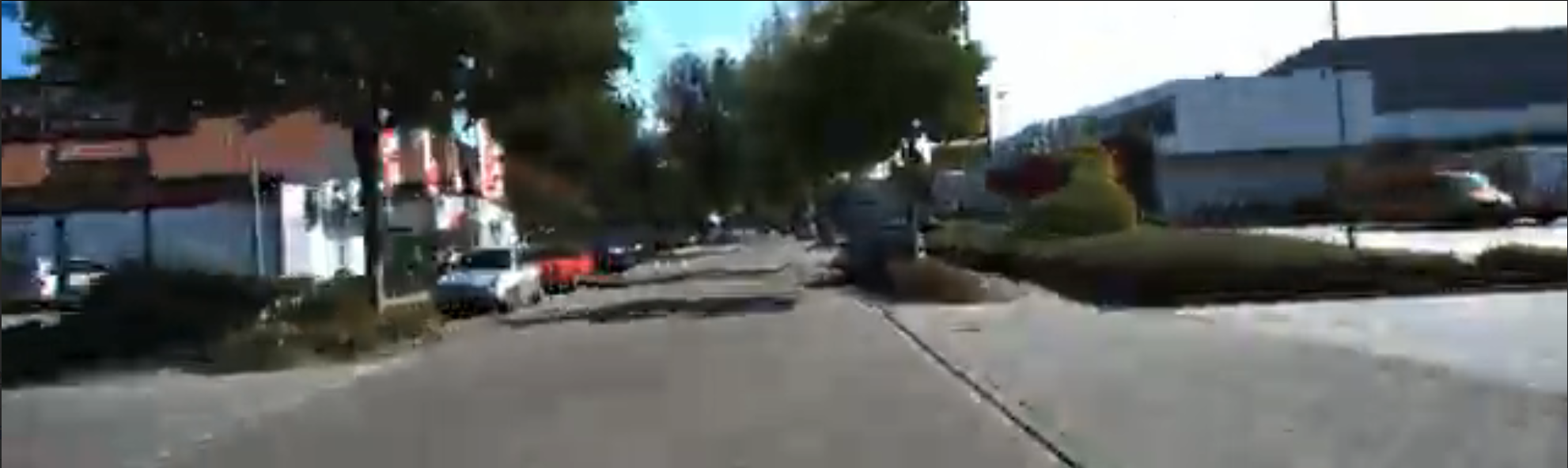} &
      \includegraphics[width=\linewidth,keepaspectratio]{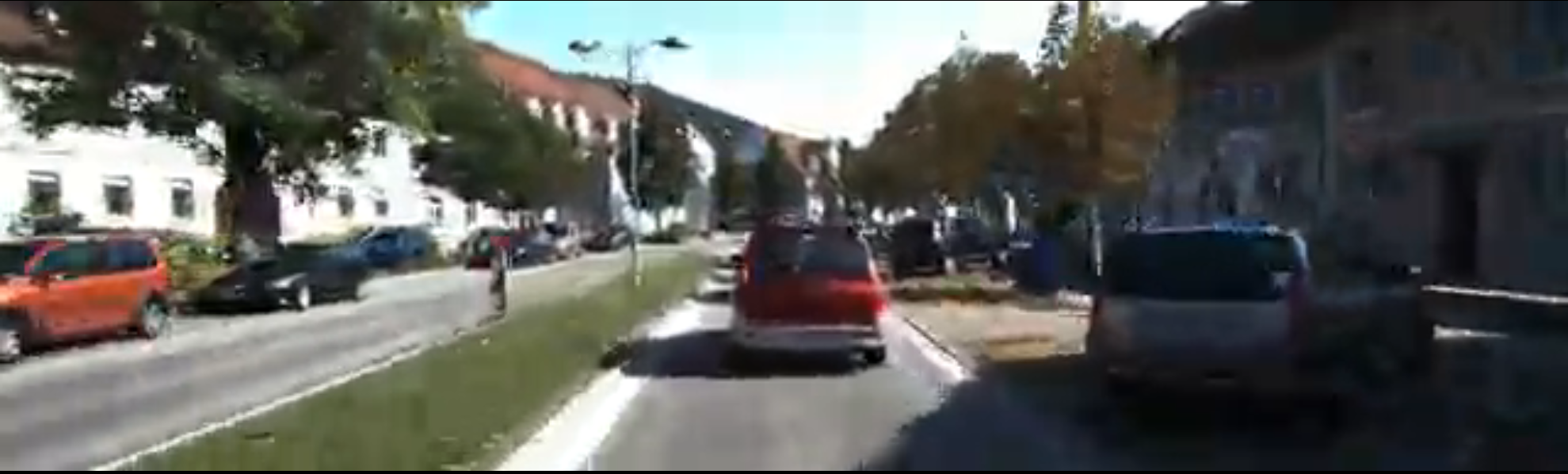} &
      The vehicle is traveling on a street with parked cars on the right side.
      There are sidewalks and grassy areas adjacent to the road. There are pedestrian crossings with signage.
      There are other vehicles moving in the same direction and parked. \\ \hline

      \includegraphics[width=\linewidth,keepaspectratio]{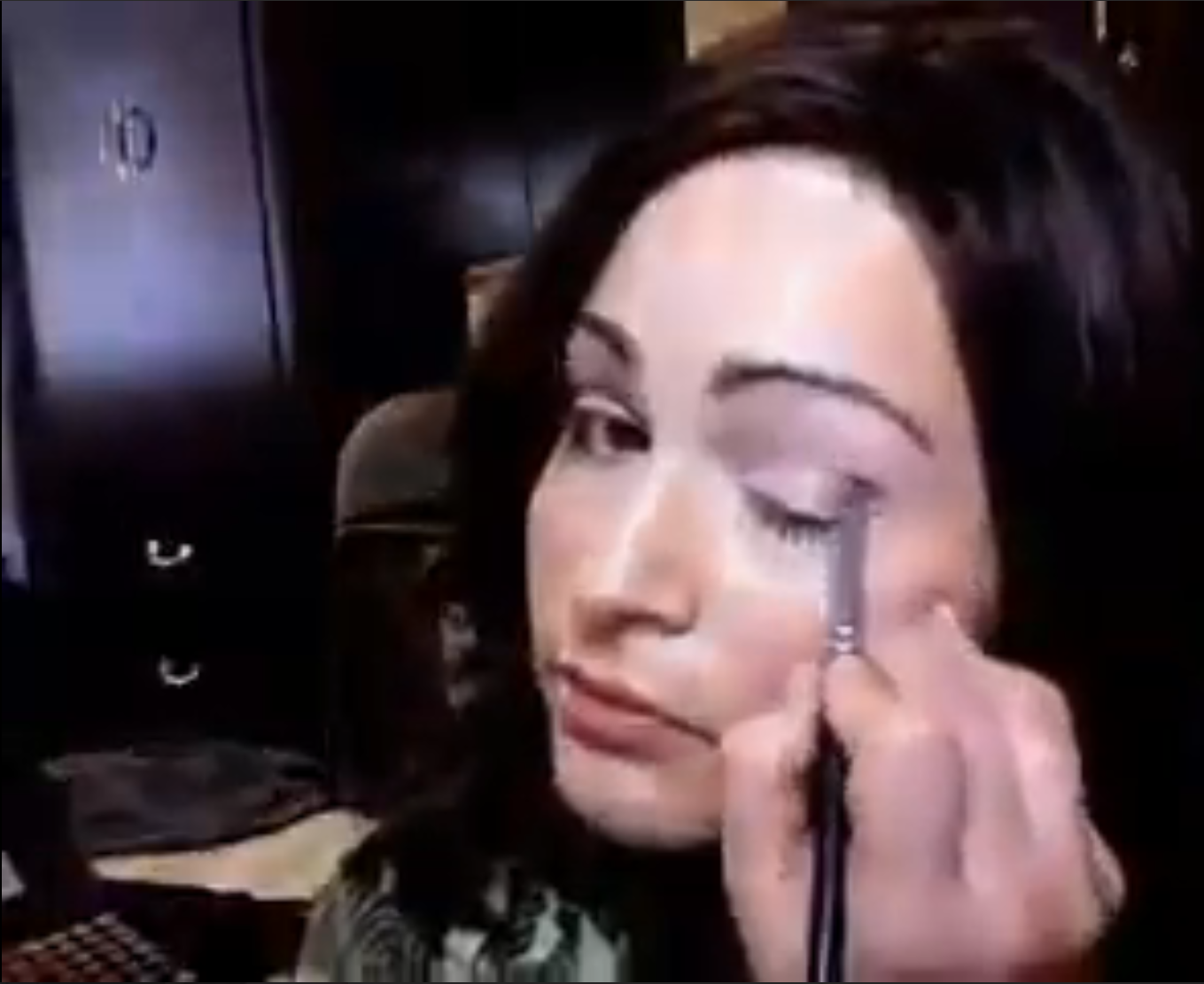} &
      \includegraphics[width=\linewidth,keepaspectratio]{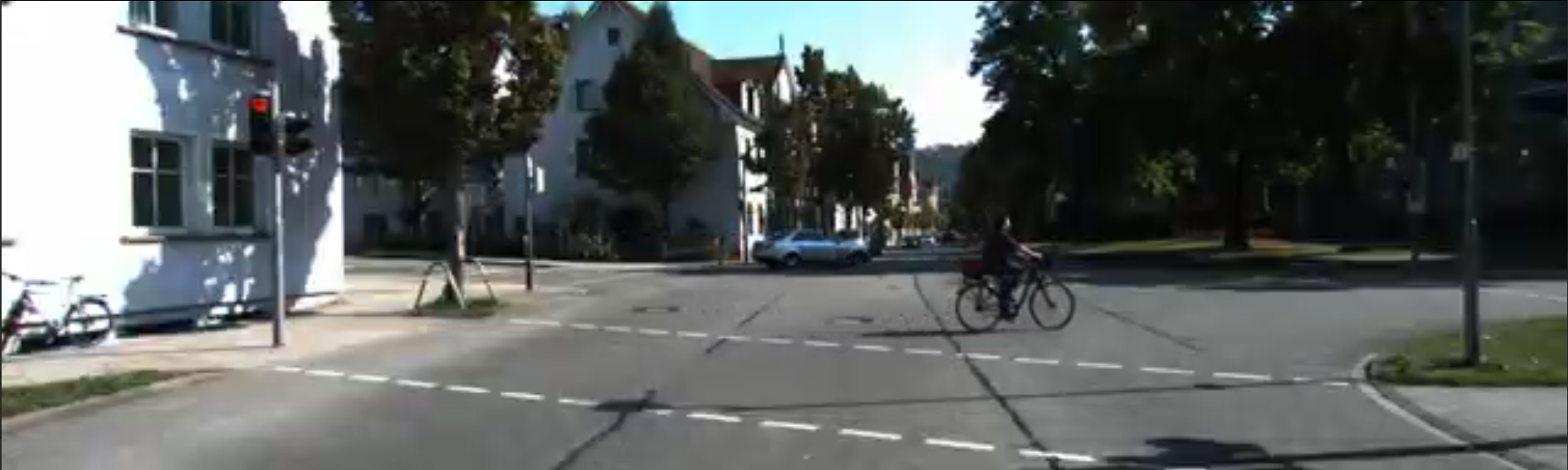} &
      The video captures a street scene with a cyclist initially. The cyclist rides across the scene from right to left.
      As the video progresses the cyclist exits the frame, and then a person walking on the sidewalk on the left side of the road becomes visible. \\ \hline

      \includegraphics[width=\linewidth,keepaspectratio]{figs/kitti_0003_eye.png} &
      \includegraphics[width=\linewidth,keepaspectratio]{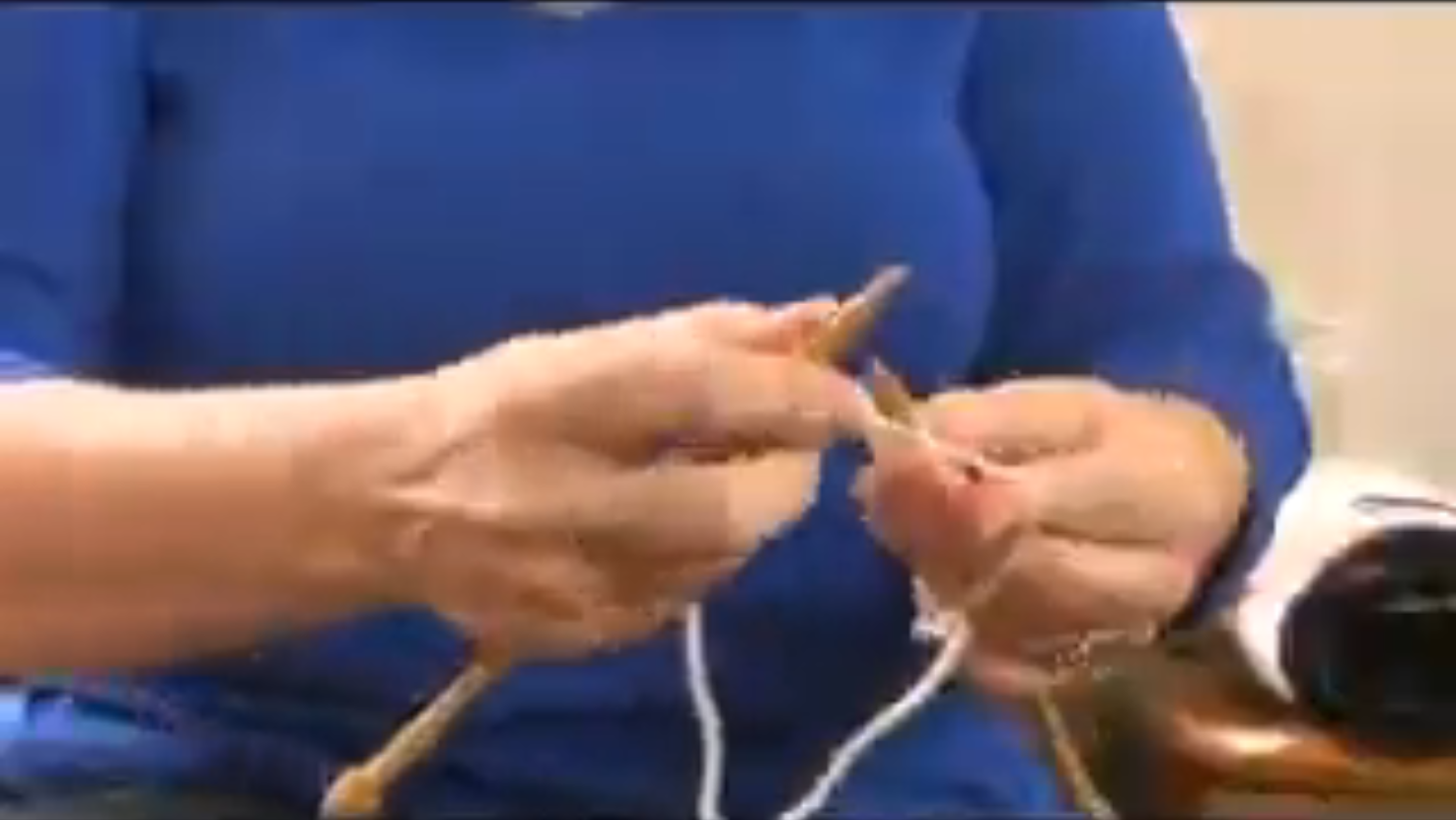} &
      The video shows a close-up view of someone knitting. The primary focus is on the hands and knitting needles as they manipulate yarn to create stitches.
      The actions appear to be those of a standard knitting technique. Due to the visual quality, it’s difficult to pinpoint the specific stitch being made, but it looks like a basic knit stitch. \\ \hline
    \end{tabular}
  \end{adjustbox}
  \label{tab:attack_performance_llm}
\end{table}

\section{Related Work}

\subsubsection{Security of Image Preprocessing Pipelines.}

In the domain of black‑box adversarial attacks, a large body of work has proposed query‑based iterative methods to improve attack efficiency and reduce reliance on substitute models, such as Square Attack~\cite{andriushchenko2020square}, Boundary Attack~\cite{brendel2017decision}, HopSkipJumpAttack~\cite{chen2020hopskipjumpattack}, GenAttack~\cite{alzantot2019genattack}, the triangle attack~\cite{moon2019parsimonious}, bandit‑based approaches~\cite{ilyas2018prior}, and SimBA~\cite{guo2019simple}. Although these methods enhance query efficiency, they still require hundreds to thousands of queries and are generally tailored to specific models. At the same time, image‑scaling attacks exploit preprocessing‑stage resizing algorithms to conceal malicious payloads within benign images. Most closely related to our work, AlphaDog~\cite{xia2025alphadog} is a `no-box' camouflage attack to exploit the alpha channel of RGBA images; it embeds the adversarial target into the RGB channels to mislead AI classifiers while crafting the alpha channel so that human observers see only innocuous content, achieving zero queries and model-agnostic applicability. In this paper, we move from the image/classification setting to the video/object-detection setting, where temporal consistency and region-level localization (rather than global labels) make attacks substantially harder. 

\subsubsection{Security of Vision-based Perception Systems.}
 
Vision is central to cyber-physical systems (e.g., AVs), but has proven vulnerable to both perturbation and patch attacks that directly corrupt input images. 

Both perturbation and patch attacks fall under the umbrella of adversarial attacks. The fundamental idea of adversarial attacks is to induce significant errors in a deep learning model’s output through minimal modifications to the input. Perturbation attacks typically affect all pixels in the input image with slight value changes, whereas patch attacks modify only a small region but with relatively large alterations in pixel values. For example, V-Phanton~\cite{10888148} introduces adversarial perturbations in captured images by adjusting the camera's supply voltage, thus disrupting the downstream image recognition process. GhostShot~\cite{ren2025ghostshot}, on the contrary, achieves the injection of adversarial patterns into CCD cameras through externally applied electromagnetic interference. Cheng et al.~\cite{cheng2023adversarial} demonstrate that the image stabilization mechanism used in autonomous driving camera sensors can be disrupted by malicious ultrasonic signals, inducing abnormal jitter and dynamic blur in acquired images. L-Hawk~\cite{liu2025hawk} adopts a similar attack concept against autonomous driving platforms, but replaces the injected signal with a laser beam precisely aimed at the camera lens. These attacks require physical access/proximity, environmental control, or specialized hardware, and often produce conditions (e.g., jitter, blur, overexposure) that can be perceptible or operationally constraining. \attack attack does not interact with the sensor or the physical environment. Instead, it targets the downstream video handling and model-input stack by exploiting how RGBA videos are decoded and consumed by perception models. 

\section{Discussion}

In this section, we discuss the practicality, limitations, and the potential defense method of the \attack attack. 

\subsection{Attack Feasibility and Practicality} 
Our work demonstrates that no-box alpha channel-based attacks can be both simple and effective. Unlike existing video attacks that rely on temporal perturbations and require knowledge of the model architecture or parameters, \attack exploits a fundamental inconsistency in video input handling. It embeds an adversarial payload into the alpha channel, which is commonly ignored or discarded by object detectors. As a result, \attack remains lightweight, broadly applicable, and agnostic to detector architectures and media playback environments.

\subsection{Limitations}

Although \attack is effective against a wide range of object detectors, it depends on two key assumptions: detectors discard the alpha channel while standard video players preserve it, conditions met by most systems trained solely on RGB inputs. Furthermore, \attack is inherently limited to grayscale content, since the alpha channel controls only pixel transparency without altering the relative intensities of the RGB channels. Consequently, the fusion mechanism cannot reproduce full‑color scenes, restricting the attack to monochrome videos or regions.

\subsection{Defense}
\attack exploits the mismatch in video pipelines in how video content is presented to human viewers compared to detection models. It leverages the removal of the alpha channel by models to hide adversarial content in plain sight. To defend against this attack, model designers can implement alpha channel profiling techniques tailored for video input during the preprocessing stage. After decoding each video frame, but before model inference, the system can conduct per-frame alpha channel analysis and compute per-pixel intensity histograms to detect any unnatural transparency distributions. Frames exhibiting nonuniform transparency in regions where transparency is not expected, such as in the center of the video in high-traffic videos, will be flagged by the detector.

Alternatively, instead of discarding the alpha channel in its entirety, a model can first composite each incoming RGBA frame onto a black background before passing it to the object detection model. This approach emulates how standard video players render transparent regions, ensuring consistency between human and model perception. Therefore, the opportunity to exploit rendering mismatches is eliminated.

\section{Conclusion}

In this paper, we introduce \attack, the first no-box adversarial attack targeting object detection systems in the video domain. By leveraging the alpha channel, we demonstrate that adversarial content can be stealthily embedded within videos without any perceptible distortion to human viewers. Our approach addresses the unique challenges of video processing by proposing multi-frame fusion. Extensive evaluations across a range of object detectors, a vision-language model, and an LLM confirm the attack's robustness and universality, achieving a 100\% success rate in all scenarios. Our findings highlight a critical and previously overlooked threat vector in cyber-physical systems, emphasizing the urgent need for new defense mechanisms to protect video-based AI applications from invisible adversarial manipulation.


\bibliographystyle{IEEEtranS}
\footnotesize
\bibliography{reference}

\begin{thebibliography}{10}
\providecommand{\url}[1]{#1}
\csname url@samestyle\endcsname
\providecommand{\newblock}{\relax}
\providecommand{\bibinfo}[2]{#2}
\providecommand{\BIBentrySTDinterwordspacing}{\spaceskip=0pt\relax}
\providecommand{\BIBentryALTinterwordstretchfactor}{4}
\providecommand{\BIBentryALTinterwordspacing}{\spaceskip=\fontdimen2\font plus
\BIBentryALTinterwordstretchfactor\fontdimen3\font minus \fontdimen4\font\relax}
\providecommand{\BIBforeignlanguage}[2]{{%
\expandafter\ifx\csname l@#1\endcsname\relax
\typeout{** WARNING: IEEEtranS.bst: No hyphenation pattern has been}%
\typeout{** loaded for the language `#1'. Using the pattern for}%
\typeout{** the default language instead.}%
\else
\language=\csname l@#1\endcsname
\fi
#2}}
\providecommand{\BIBdecl}{\relax}
\BIBdecl

\bibitem{alzantot2019genattack}
M.~Alzantot, Y.~Sharma, S.~Chakraborty, H.~Zhang, C.-J. Hsieh, and M.~B. Srivastava, ``Genattack: Practical black-box attacks with gradient-free optimization,'' in \emph{Proceedings of the genetic and evolutionary computation conference}, 2019, pp. 1111--1119.

\bibitem{andriushchenko2020square}
M.~Andriushchenko, F.~Croce, N.~Flammarion, and M.~Hein, ``Square attack: a query-efficient black-box adversarial attack via random search,'' in \emph{European conference on computer vision}.\hskip 1em plus 0.5em minus 0.4em\relax Springer, 2020, pp. 484--501.

\bibitem{brendel2017decision}
W.~Brendel, J.~Rauber, and M.~Bethge, ``Decision-based adversarial attacks: Reliable attacks against black-box machine learning models,'' \emph{arXiv preprint arXiv:1712.04248}, 2017.

\bibitem{chen2020hopskipjumpattack}
J.~Chen, M.~I. Jordan, and M.~J. Wainwright, ``Hopskipjumpattack: A query-efficient decision-based attack,'' in \emph{2020 ieee symposium on security and privacy (sp)}.\hskip 1em plus 0.5em minus 0.4em\relax IEEE, 2020, pp. 1277--1294.

\bibitem{cheng2023adversarial}
Y.~Cheng, X.~Ji, W.~Zhu, S.~Zhang, K.~Fu, and W.~Xu, ``Adversarial computer vision via acoustic manipulation of camera sensors,'' \emph{IEEE Transactions on Dependable and Secure Computing}, vol.~21, no.~4, pp. 3734--3750, 2023.

\bibitem{geiger2013vision}
A.~Geiger, P.~Lenz, C.~Stiller, and R.~Urtasun, ``Vision meets robotics: The kitti dataset,'' \emph{The international journal of robotics research}, vol.~32, no.~11, pp. 1231--1237, 2013.

\bibitem{guo2019simple}
C.~Guo, J.~Gardner, Y.~You, A.~G. Wilson, and K.~Weinberger, ``Simple black-box adversarial attacks,'' in \emph{International conference on machine learning}.\hskip 1em plus 0.5em minus 0.4em\relax PMLR, 2019, pp. 2484--2493.

\bibitem{ilyas2018prior}
A.~Ilyas, L.~Engstrom, and A.~Madry, ``Prior convictions: Black-box adversarial attacks with bandits and priors,'' \emph{arXiv preprint arXiv:1807.07978}, 2018.

\bibitem{10888148}
Y.~Jiang, R.~Li, Y.~Cheng, X.~Ji, and W.~Xu, ``V-phanton:voltage-based physically-triggered backdoor attack against facial recognition,'' in \emph{ICASSP 2025 - 2025 IEEE International Conference on Acoustics, Speech and Signal Processing (ICASSP)}, 2025, pp. 1--5.

\bibitem{Jocher_YOLOv5_by_Ultralytics_2020}
\BIBentryALTinterwordspacing
G.~Jocher, ``{YOLOv5 by Ultralytics},'' May 2020. [Online]. Available: \url{https://github.com/ultralytics/yolov5}
\BIBentrySTDinterwordspacing

\bibitem{yolov8_ultralytics}
\BIBentryALTinterwordspacing
G.~Jocher, A.~Chaurasia, and J.~Qiu, ``Ultralytics yolov8,'' 2023. [Online]. Available: \url{https://github.com/ultralytics/ultralytics}
\BIBentrySTDinterwordspacing

\bibitem{yolo11_ultralytics}
\BIBentryALTinterwordspacing
G.~Jocher and J.~Qiu, ``Ultralytics yolo11,'' 2024. [Online]. Available: \url{https://github.com/ultralytics/ultralytics}
\BIBentrySTDinterwordspacing

\bibitem{khanam2024yolov5deeplookinternal}
\BIBentryALTinterwordspacing
R.~Khanam and M.~Hussain, ``What is yolov5: A deep look into the internal features of the popular object detector,'' 2024. [Online]. Available: \url{https://arxiv.org/abs/2407.20892}
\BIBentrySTDinterwordspacing

\bibitem{lin2018focallossdenseobject}
\BIBentryALTinterwordspacing
T.-Y. Lin, P.~Goyal, R.~Girshick, K.~He, and P.~Dollár, ``Focal loss for dense object detection,'' 2018. [Online]. Available: \url{https://arxiv.org/abs/1708.02002}
\BIBentrySTDinterwordspacing

\bibitem{liu2025hawk}
T.~Liu, Y.~Liu, Z.~Ma, T.~Yang, X.~Liu, T.~Li, and J.~Ma, ``L-hawk: A controllable physical adversarial patch against a long-distance target.'' in \emph{NDSS}, 2025.

\bibitem{moon2019parsimonious}
S.~Moon, G.~An, and H.~O. Song, ``Parsimonious black-box adversarial attacks via efficient combinatorial optimization,'' in \emph{International conference on machine learning}.\hskip 1em plus 0.5em minus 0.4em\relax PMLR, 2019, pp. 4636--4645.

\bibitem{ren2016faster}
S.~Ren, K.~He, R.~Girshick, and J.~Sun, ``Faster r-cnn: Towards real-time object detection with region proposal networks,'' \emph{IEEE transactions on pattern analysis and machine intelligence}, vol.~39, no.~6, pp. 1137--1149, 2016.

\bibitem{ren2025ghostshot}
Y.~Ren, Q.~Jiang, C.~Yan, X.~Ji, and W.~Xu, ``Ghostshot: Manipulating the image of ccd cameras with electromagnetic interference.''

\bibitem{soomro2012ucf101}
K.~Soomro, A.~R. Zamir, and M.~Shah, ``Ucf101: A dataset of 101 human actions classes from videos in the wild,'' \emph{arXiv preprint arXiv:1212.0402}, 2012.

\bibitem{Suri2016}
R.~Suri, ``Is it okay to convert an image with an alpha channel to an image without an alpha channel?'' \url{https://stackoverflow.com/questions/37877020/is-it-okay-to-convert-an-image-with-an-alpha-channel-to-an-image-without-an-alph}, Jun. 2016, [Online; accessed 30-July-2025].

\bibitem{team2023gemini}
G.~Team, R.~Anil, S.~Borgeaud, J.-B. Alayrac, J.~Yu, R.~Soricut, J.~Schalkwyk, A.~M. Dai, A.~Hauth, K.~Millican \emph{et~al.}, ``Gemini: a family of highly capable multimodal models,'' \emph{arXiv preprint arXiv:2312.11805}, 2023.

\bibitem{starship}
S.~Technologies, ``Starship: Our robots,'' \url{https://www.starship.xyz/our-robots/}, 2014, accessed: 2025-07-25.

\bibitem{tesla_fsd}
Tesla, ``Autopilot and full self-driving (supervised),'' \url{https://www.tesla.com/support/autopilot}, 2022, accessed: 2025-07-25.

\bibitem{wang2023beyond}
G.~Wang, C.~Zhou, Y.~Wang, B.~Chen, H.~Guo, and Q.~Yan, ``Beyond boundaries: A comprehensive survey of transferable attacks on ai systems,'' \emph{arXiv preprint arXiv:2311.11796}, 2023.

\bibitem{weng2023open}
Z.~Weng, X.~Yang, A.~Li, Z.~Wu, and Y.-G. Jiang, ``Open-vclip: Transforming clip to an open-vocabulary video model via interpolated weight optimization,'' in \emph{International conference on machine learning}.\hskip 1em plus 0.5em minus 0.4em\relax PMLR, 2023, pp. 36\,978--36\,989.

\bibitem{xia2025alphadog}
Q.~Xia and Q.~Chen, ``Alphadog: No-box camouflage attacks via alpha channel oversight.'' in \emph{NDSS}, 2025.

\bibitem{zhou2024comprehensive}
C.~Zhou, Q.~Li, C.~Li, J.~Yu, Y.~Liu, G.~Wang, K.~Zhang, C.~Ji, Q.~Yan, L.~He \emph{et~al.}, ``A comprehensive survey on pretrained foundation models: A history from bert to chatgpt,'' \emph{International Journal of Machine Learning and Cybernetics}, pp. 1--65, 2024.

\bibitem{zhou2024optical}
C.~Zhou, Q.~Yan, D.~Kent, G.~Wang, Z.~Zhang, and H.~Radha, ``Optical lens attack on deep learning based monocular depth estimation,'' \emph{arXiv preprint arXiv:2409.17376}, 2024.

\bibitem{zhou2022doublestar}
C.~Zhou, Q.~Yan, Y.~Shi, and L.~Sun, ``Doublestar: Long-range attack towards depth estimation based obstacle avoidance in autonomous systems,'' in \emph{31st USENIX security symposium (USENIX Security 22)}, 2022, pp. 1885--1902.

\bibitem{zou2023object}
Z.~Zou, K.~Chen, Z.~Shi, Y.~Guo, and J.~Ye, ``Object detection in 20 years: A survey,'' \emph{Proceedings of the IEEE}, vol. 111, no.~3, pp. 257--276, 2023.

\end{thebibliography}

\end{document}